\documentclass[10pt,letterpaper,twoside,twocolumn,journal,final]{IEEEtran}
\usepackage[T1]{fontenc}
\usepackage{gensymb,graphicx,booktabs}
\usepackage[noadjust]{cite}
\usepackage[ruled,vlined,linesnumbered]{algorithm2e}
\usepackage[abs]{overpic}
\usepackage{textcomp,latexsym,tikz,array}
\usepackage{booktabs,threeparttable,lineno}
\usepackage{amsfonts,amsmath,mathtools}
\usepackage{hyperref}
\usetikzlibrary{matrix}
\graphicspath{{figs/}}
\newcommand{\graphheightscale}{0.152}

\makeatletter
\newcommand{\removelatexerror}{\let\@latex@error\@gobble}
\makeatother

\begin{document}

\title{Evolutionary n-level Hypergraph Partitioning with Adaptive Coarsening}
\author{Richard~J.~Preen and Jim~Smith%
	\thanks{Manuscript date of current version October 5, 2018.}%
	\thanks{The authors are with the Department of Computer Science and Creative Technologies, University of the West of England, Bristol BS16 1QY, UK (e-mail: richard2.preen@uwe.ac.uk; james.smith@uwe.ac.uk).}%
	\thanks{Digital Object Identifier 10.1109/TEVC.2019.2896951}%
}

\bstctlcite{IEEEexample:BSTcontrol}
\IEEEpubid{}
\markboth{IEEE TRANSACTIONS ON EVOLUTIONARY COMPUTATION, DOI:10.1109/TEVC.2019.2896951}%
{PREEN \MakeLowercase{and} SMITH:~EVOLUTIONARY n-LEVEL HYPERGRAPH PARTITIONING WITH ADAPTIVE COARSENING}

\maketitle

\begin{abstract}
Hypergraph partitioning is an NP-hard problem that occurs in many computer science applications where it is necessary to reduce large problems into a number of smaller, computationally tractable sub-problems. Current techniques use a multilevel approach wherein an initial partitioning is performed after compressing the hypergraph to a predetermined level. This level is typically chosen to produce very coarse hypergraphs in which heuristic algorithms are fast and effective. This article presents a novel memetic algorithm which remains effective on larger initial hypergraphs. This enables the exploitation of information that can be lost during coarsening and results in improved final solution quality. We use this algorithm to present an empirical analysis of the space of possible initial hypergraphs in terms of its searchability at different levels of coarsening. We find that the best results arise at coarsening levels unique to each hypergraph. Based on this, we introduce an adaptive scheme that stops coarsening when the rate of information loss in a hypergraph becomes non-linear and show that this produces further improvements. The results show that we have identified a valuable role for evolutionary algorithms within the current state-of-the-art hypergraph partitioning framework.
\end{abstract}


\IEEEpeerreviewmaketitle

\section{Introduction}

\IEEEPARstart{H}{ypergraph partitioning} (HGP) is an NP-hard problem~\cite{Lengauer:1990} that occurs in many computer science applications where it is necessary to reduce large problems into a number of smaller, computationally tractable sub-problems. Common applications include very large scale integration (VLSI) design~\cite{Areibi:2004} and scientific computing~\cite{Selvitopi:2017}. 

Hypergraphs are a generalisation of graphs where each hyperedge may connect more than two vertices. Formally, a hypergraph can be defined ~\cite{Trifunovic:2006,Lotfifar:2016}  as $H = \{V,\mathcal{E},c,\omega\}$ where:
\begin{itemize}
    \item 
$V=\{v_1,\ldots,v_n\}$  and  $\mathcal{E}=\{e_1,\ldots,e_m\}$ are  finite sets of vertices and hyperedges. 
\item Edges and vertices may have  associated weights: $c(v)$ denotes the weight of a vertex $v \in V$ and  $\omega(e)$ denotes the weight of a hyperedge $e \in \mathcal{E}$.
\end{itemize}
A hyperedge $e \in \mathcal{E}$ is said to be incident on a vertex $v \in V$  if, and only if, $v \in e$. Vertices $u,v \in V$ are said to be adjacent in a hypergraph, if, and only if, there exists a hyperedge $e \in \mathcal{E}$ such that $u \in e$ and $v \in e$. The degree of a vertex $d(v)$ is the number of distinct hyperedges in $\mathcal{E}$ that are incident on $v$, and the length of a hyperedge is defined as its cardinality $|e|$.

\begin{figure}[t]
    \removelatexerror
    \begin{algorithm}[H]
    	\SetNoFillComment
    	\small
    	\SetAlgoLined
    	\DontPrintSemicolon
    	{\bf Input:} Hypergraph $H(V,\mathcal{E},c,\omega)$, $t$, $k$, $\epsilon$\;
    	\tcc{coarsening}
    	\While{$|V'| > k \times t$} {
    		$\{a,b\} = SelectNodes(V')$\;
            $V' \leftarrow Contract(V',a,b)$\;
        }
        $PartitionList_{V'} = InitialPartition(V',k,\epsilon)$\;
        \tcc{uncoarsening}
        \While{$|V'| < |V|$} {
            $c \leftarrow SelectNodeToExpand(V')$\;
            $V' \leftarrow Expand(V',c)$\;
            $PartitionList_{V'} \leftarrow Refine(V', PartitionList_{V'}) $\;
        }
    	{\bf Output:} $PartitionList_V$
    	\caption{Multilevel Hypergraph Partitioning}
    	\label{alg:multilevel}
    \end{algorithm}
\end{figure}

The $k$-way HGP problem is to partition the set of vertices into $k$ approximately equal disjoint subsets whilst minimising an objective function. Typically this is the {\em cut-size}: the sum of the weights of those hyperedges that span different subsets. However, minimising cut-size often leads to an uneven distribution of the cut hyperedges between partitions. Alternatives are the sum of external degrees, and $(K-1)$ metric, which includes the number of subsets connected by a hyperedge~\cite{Lotfifar:2016}.

\IEEEpubidadjcol

Current state-of-the-art algorithms, including MLPart~\cite{Alpert:1998}, hMetis~\cite{Karypis:1999}, PaToH~\cite{Catalyurek:1999}, Zoltan~\cite{Devine:2006}, Par$k$way~\cite{Trifunovic:2008}, UMPa~\cite{Catalyurek:2013}, and KaHyPar~\cite{Schlag:2016}, use a multilevel approach as illustrated in Algorithm~\ref{alg:multilevel}. The approach recursively coarsens a hypergraph by contracting a single pair of vertices at each level until $t \times k$ hypernodes remain. During coarsening, KaHyPar, hMetis, and PaToH use a greedy heavy-edge rating function in $SelectNodes()$ however more sophisticated techniques respecting the community structure have recently been explored~\cite{Heuer:2017}. Various methods may be used to generate the assignment of super-nodes to partitions in $InitialPartition()$. This assignment is further improved using the Fiduccia-Mattheyses~\cite{Fiduccia:1982} (FM) move-based local search algorithm. The uncoarsening phase recursively selects a node to expand (e.g., $\{a,b\} \leftarrow c$) and then uses FM to refine which partition nodes $a$ and $b$ are assigned. Using a larger number of levels~\cite{Osipov:2010} and performing repeated iterations of the entire multilevel partitioning, known as V-cycles~\cite{Karypis:1999}, can improve the solution quality, albeit at a computational cost. 

Direct $k$-way partitioning (Algorithm~\ref{alg:multilevel}) has the potential advantage of allowing the search algorithm to take a global view. This can result in better solutions for large hypergraphs and tighter balance constraints~\cite{Akhremtsev:2017}. However, for scalability reasons recursive bisection approaches are more widely used.

Despite their sophistication, it is notable that these approaches stop coarsening at some predefined threshold $t$ of remaining supernodes. Most implementations, such as hMetis, PaToH, and KaHyPar, use default thresholds of $t \approx 150$, resulting in hypergraphs with around 300 vertices for initial partitioning. This value may result in fast and reasonably effective heuristic algorithms, but does not necessarily correspond to a good trade-off between scale and information content.

Karypis and Kumar~\cite{Karypis:1998} showed that a good partitioning of the coarsest hypergraph \emph{generally} leads to a good partitioning of the original hypergraph. This can reduce the amount of time spent on refinement in the uncoarsening phase. However, it is important to note that the initial hypergraph partitioning with the smallest cut-size may not necessarily lead to the smallest final cut-size after refinement is performed during uncoarsening~\cite{Karypis:2003}. Since information may be hidden to the global optimisation algorithm during compression, the more the hypergraph is coarsened the greater this effect may be.

Many approaches have been developed to perform the initial partitioning, ranging from random assignment~\cite{Alpert:1998} to the use of various greedy growing techniques~\cite{Catalyurek:1999}, recursive bisection~\cite{Karypis:1999}, and evolutionary algorithms (EAs)~\cite{Soper:2004}. Greedy growth algorithms quickly produce balanced partitions, but are sensitive to the initial randomly chosen vertex~\cite{Catalyurek:1999}. Since the initial partitioning usually takes place on very small hypergraphs these algorithms can be rerun multiple times. The best partitioning found is subsequently propagated for refinement during the uncoarsening phase~\cite{Catalyurek:1999}.

It is difficult to generalise measures to select the optimal algorithm to use for a given problem instance, i.e., the algorithm selection problem~\cite{Kotthoff:2014}. Therefore, a portfolio approach is used in practice by PaToH, hMetis, and KaHyPar~\cite{Heuer:2015}. For example, PaToH uses 11 different random and greedy growth heuristic algorithms~\cite{PaToH}. The KaHyPar `Pool' portfolio approach to initial partitioning also uses a range of simple algorithms, including fully random, breadth-first search (BFS), label propagation, and nine variants of greedy hypergraph growing. Each algorithm is executed $r$ number of times, then the partition with the smallest cut-size and lowest imbalance is presented for uncoarsening where it is projected back to the original hypergraph. This approach has been extensively parameter tuned~\cite{Heuer:2015}, finding that $r$ = 20 produces the overall best results at $t$ = 150, with partitions that are only marginally worse than $r$ = 75, yet significantly faster. Over a wide range of hypergraphs this approach has recently been shown to identify similar or better partitions in a faster time than the most popular general purpose HGP algorithms, hMetis and PaToH~\cite{Schlag:2016,Akhremtsev:2017}, neither of which are open source.

In this article, we examine the case where there exists a large computational budget and many evaluations can be performed on less coarsened hypergraphs to identify the best final partitions, i.e., the potential for larger $r$ and $t$ exists. We explore the use of EAs to perform the initial partitioning within the state-of-the-art, open source (GPLv3), Karlsruhe $n$-level hypergraph partitioning framework, KaHyPar from \url{https://github.com/SebastianSchlag/kahypar}.

In particular, the following contributions are made:
\begin{enumerate}
    \item We characterise the `searchability' of the space of initial partitions   at different levels of coarsening.
    \item Based on that analysis, we identify a role for EAs in terms of the level of coarsening, and hence the speed vs.\@ quality of solutions produced. We also identify some key algorithm characteristics.
    \item We develop a novel memetic algorithm and demonstrate that this discovers significantly better final solutions across a range of classes of hypergraphs and across a range of different coarsening thresholds.
    \item Finally, we develop an adaptive mechanism for deciding when to perform initial partitioning based on the rate of change of information content in the hypergraph as it is coarsened. We show that this also gives significant performance improvements.
\end{enumerate}

In the remainder of this article, Section~\ref{sec:background} discusses the related work. Section~\ref{sec:methodology} describes the test framework, the memetic-EA initial partitioner, and comparison metrics. Section~\ref{sec:landscape} presents a landscape analysis with respect to EA design at different levels of coarsening. Section~\ref{sec:sensitivity} presents the results of parameter sensitivity testing. Section~\ref{sec:adaptive} introduces and presents results from a novel adaptive coarsening algorithm to identify the EA niche. Finally, Section~\ref{sec:conclusions} summarises the conclusions.

\section{Related Work}
\label{sec:background}

Many EAs have been applied to the more well-known problem of graph partitioning; see Kim~{\it et~al.}~\cite{Kim:2011} for an overview. Soper~{\it et~al.}~\cite{Soper:2004} were the first to use an EA within a multilevel approach. They introduced variation operators that modify the edge weights of the graph depending on the input partitions. Subsequently presenting these to a multilevel partitioner, which uses the weights to obtain a new partition.

More recently, Benlic and Hao~\cite{Benlic:2011} used a memetic algorithm within a multilevel approach to solve the perfectly balanced graph partitioning problem $\epsilon=0$. They hypothesised that a large number of vertices will always be grouped together among high quality partitions and introduced a multiparent crossover operator, with the offspring being refined with a perturbation-based tabu search algorithm. 

Sanders and Schulz~\cite{Sanders:2012} used an EA within a multilevel approach and showed that the usage of edge weight perturbations decreases the overall quality of the underlying graph partitioner; subsequently introducing new crossover and mutation operators that avoid randomly perturbing the edge weights. Their algorithm has recently been incorporated within a faster parallelised approach~\cite{Meyerhenke:2017}.

In addition to performing the initial partitioning, EAs can also be used in other areas of the multilevel approach. For example, K\"{u}\c{c}\"{u}kpetek~{\it et~al.}~\cite{Kucukpetek:2005} used an EA to perform the coarsening phase in a multilevel graph partitioning algorithm.

Merz and Freisleben~\cite{Merz:2000} showed that the fitness landscape depends on the structure of the graph and, perhaps unintuitively, that the landscape can become smoother as the average degree increases. Consequently, Pope~{\it et~al.}~\cite{Pope:2016} proposed the use of genetic programming as a meta-level algorithm to select the best combination of existing algorithms for coarsening, partitioning, and refinement, based on the characteristics of the graph being solved.
 
The most popular chromosome representation is group-number encoding, wherein each gene represents the partition group to assign a given vertex, i.e., there are as many genes as there are vertices $|V|$ and alleles as there are partitions. This has led to a wide variety of proposed crossover and normalisation schemes since different assignments of allele values to groups still represent the same solution. For example, M\"{u}hlenbein and Mahnig~\cite{Muhlenbein:2002} used the simple normalisation technique of inverting each candidate and selecting the one with the smallest Hamming distance. 


EAs have been relatively under-explored for the more general case of HGP however: there has been a small amount of prior work on VLSI circuit partitioning. For example, Schwarz and O\u{c}en\'{a}\u{s}ek~\cite{Schwarz:1999} briefly studied several EAs including the Bayesian optimisation algorithm for direct (i.e., not multilevel) small VLSI partitioning. Kim~{\it et~al.}~\cite{Kim:2004} explored a memetic algorithm using a modified FM for local optimisation and reported smaller bipartition cut-sizes on a number of benchmark circuits when compared with hMetis. Notably, Areibi and Yang~\cite{Areibi:2004} explored VLSI design via the use of memetic algorithms using FM for local optimisation within a multilevel approach and reported improvements of 35\% over a simple genetic algorithm. This has since been implemented in hardware using reconfigurable computing~\cite{Coe:2007}. Significantly, none of these algorithms are considered to be competitive with state-of-the-art hypergraph partitioning tools.

Recently a memetic EA has been introduced to build on the KaHyPar framework~\cite{Andre:2017}. This algorithm runs a steady-state EA with a population at the original uncoarsened level. The initial population is seeded using a variant of KaHyPar. Each generation, binary tournament selection is used to choose two parents, then variation operators are applied to the fitter of those, running a number of V-cycles of coarsening--initial partitioning--uncoarsening, using different randomisation seeds. The recombination operator only runs V-cycles on the subset of original-level vertices that are in different partitions in the two parents. Two mutation operators were defined: one starting from the original level, and another which preserves more locality by skipping the coarsening phase and starting from the initial partition corresponding to the fitter parent (these are cached to save time.) To maintain diversity, a variant of restricted tournament selection is used and the authors introduce a novel distance measure that they claim is better suited to this problem domain than Hamming distance.

The work presented here and that in \cite{Andre:2017} share the idea that the memetic algorithm should work at a less coarsened level. However, there are key differences: in \cite{Andre:2017} the EA works at the wholly uncoarsened level, which can mean millions of vertices/genes. Therefore, to make the search tractable the sub-space in which search occurs (via the V-cycles) is restricted and initial partitioning run at a highly coarsened level.

\section{Methodology}
\label{sec:methodology}

\subsection{Test Framework}

To ensure the comparability of results we use the KaHyPar $n$-level hypergraph partitioner~\cite{Heuer:2015,Schlag:2016,Akhremtsev:2017}. This is a mature toolkit to which considerable attention has been paid to parameter tuning, so no further optimisation was applied. We also use a selection of the hypergraphs used previously for benchmarking KaHyPar, available from \url{http://doi.org/10.5281/zenodo.30176}.  
Specifically, we use: the 10 largest from the well-known ISPD98 VLSI circuits~\cite{ISPD98}; and 10 each randomly selected from the University of Florida sparse matrix collection (SPM)~\cite{UF:SM} ({\it Airfoil\_2d, Reuters911, usroads, stokes128, Andrews, Baumann, HTC\_336\_9129, NotreDame\_actors, Stanford, nasasrb}) and the 2014 international SAT competition (SAT)~\cite{SAT14} ({\it gss-20-s100, MD5-28-2, ctl\_4291\_567\_5\_unsat\_pre, aaai10-planning-ipc5-pathways-17-step21, slp-synthesis-aes-top29, hwmcc10-timeframe-expansion-k45-pdtvisns, dated-10-11-u, atco\_enc1\_opt2\_05\_4, UCG-15-10p1, openstacks-p30\_3.085}).

Since KaHyPar is currently the best general state-of-the-art hypergraph partitioner~\cite{Schlag:2016,Akhremtsev:2017}, and recursive bipartitioning can scale with increasing $k$ more effectively, here we use an initial testing regime of $k$ = 2 and $\epsilon$ = 0.1. For benchmark comparisons, we use the KaHyPar Pool portfolio algorithm described above,  and compare results at equivalent numbers of evaluations. An evaluation consists of generating an initial partitioning followed by an application of the FM algorithm. However, it should be noted that one evaluation of an algorithm in the Pool (e.g., a BFS) has a longer wall-clock time than an EA evaluation. The total partitioning times for the experiments reported here are approximately $1.9\times$ longer for the Pool when compared at the same $t$ threshold. For $k$ = 2, the ($K$-1) and hyperedge cut-size metrics are identical~\cite{Trifunovic:2006}, and so here we use this as the objective function.

\subsection{Representation, Algorithm Operators and Parameters} 

We adopt a simple vertex-to-cluster encoding of the $N$ coarsened hypernodes, and use a ($\mu$+$\lambda$) EA where each subsequent generation consists of the $\mu$ fittest from the parental population and $\lambda$ offspring. Each offspring is created as the product of two (independently) randomly selected parents. Uniform crossover is applied with $\mathcal{X}$ = 80\% probability. Symmetry in the fitness landscape can severely obstruct the evolutionary search~\cite{Choi:2007}, so we apply parental alignment (normalisation) during crossover: if the Hamming distance between the parents exceeds $N/2$ then the gene values of one parent are inverted. A self-adaptive mutation scheme is then applied, setting genes to random values. Following Serpell and Smith~\cite{Serpell:2010}, each candidate maintains its own mutation rate. This is initially inherited from the fitter of its parents, and then with $\mathcal{A}$ = 10\% probability may be randomly reset to one of 10 possible values before applying mutation at the resulting rate. If an offspring has an imbalance greater than $\epsilon$, a repair mechanism is invoked, randomly moving vertices from the largest to the smallest partition. Lamarkian evolution is performed by subsequently applying the FM local search algorithm using default~\cite{Schlag:2016} KaHyPar settings and the offspring acquiring any modifications. See Algorithm~\ref{alg:ea}.

\begin{figure}[t]
    \removelatexerror
    \begin{algorithm}[H]
    	\SetNoFillComment
    	\small
    	\SetAlgoLined
    	\DontPrintSemicolon
    	$n=\frac{1}{|V|}$;
    	$\mathcal{M} = [\frac{n}{100},\frac{n}{10},\frac{n}{5},\frac{n}{2},n,n,2n,5n,10n,100n]$\;
    	initialise parent population: $P=\{a_1 \ldots a_{\mu}\}$\;
    	\While{evaluation budget not exhausted} {
    		\tcc{create offspring population}
    		\For{$i=1$ to $\lambda$} {
    			parent $p_1 = RandomSelection(P)$\;
    			parent $p_2 = RandomSelection(P)$\;
    			offspring $a_i = p_{1}$\;
    			\If{rand() $<\mathcal{X}$} {
    				perform uniform crossover with normalised $p_2$\;
    				\If{$p_1.fitness < p_2.fitness$} {
    					$a_i.mut = p_{2}.mut$\;
    				}
    			}
    			\If{rand() $<\mathcal{A}$} {
    				$a_i.mut = RandomSelection(\mathcal{M})$\;
    			}
    			\For{each hypernode in $a_i$} {
    				\If{drand() $<a_i.mut$} {
    					assign hypernode to a random partition\;
    				}
    			}
    			repair partition if necessary\;
    			apply FM local search (Lamarkian)\;
    			evaluate $a_i$\;
    		}
    		\tcc{select next parental population}
    		$P=$ $\mu$ fittest from $P+\lambda$\;
    	}
    	\caption{Memetic EA($\mu+\lambda$) initial partitioner}
    	\label{alg:ea}
    \end{algorithm}
\end{figure}

\subsection{Comparison Metrics and Statistical Analysis of Results}

The distribution of values observed from repeated runs was not normally distributed---especially when there is a `hard' lower or upper limit. We therefore apply non-parametric tests.

For each run, we recorded two values: the {\em initial} cut-size as the value found by a search algorithm operating at the coarsest level, and the {\em final} cut-size as the value at the original level, i.e., after uncoarsening has taken place. Since these values will depend on the coarsening threshold $t$ and choice of algorithm, we denote these as $cut^t_{alg}$. In some cases below we also report the {\em best-case} cut-size: $cut^*_{alg}$, the value observed at whichever coarsening threshold gave the best results for a given dataset.

To measure the performance of different algorithms across the full range of thresholds, we also present the area under the curve (AUC) results, estimated from the experiments at individual thresholds using a composite Simpson's rule. When comparing methods on a single problem, we use the Wilcoxon ranked-sums test, with the null hypothesis that all observed results come from the same distribution.

To draw any firm overall conclusions about the performance of the two approaches, we follow the recommendations in \cite{Demsar:2006} for comparing algorithms over multiple data sets. First, we examine the  results to ensure that for each algorithm-hypergraph combination the arithmetic mean is a reliable estimate of performance, i.e., that the distribution of observations from the 20 runs is unimodal with low standard deviation. This results in a pair of values (one per algorithm) for each hypergraph, to which the Wilcoxon signed ranks test can be applied with the null hypothesis that taken across all hypergraphs there is no difference in performance.

Finally, run-times are recorded as total-wall-clock time for the whole process because the time taken in each phase is heavily linked to the results of the previous stage. 

\section{Landscape Analysis at Different Levels}
\label{sec:landscape}

One of the tenets of the multilevel approach to solving HGP is that the sheer size of the search space makes it impractical to solve at the original, uncoarsened level, and that therefore it is better to conduct the search for a good initial partitioning within a much smaller space. It has also been suggested that the graph-partitioning counterparts become easier to search as the level of coarsening increases~\cite{Merz:2000}. Nevertheless, there is clearly a trade-off. It is inevitable that the coarsening process reduces the information content, so the mapping between quality of initial and final cuts becomes more noisy---especially given the greedy uncoarsening process.

To investigate the nature of the search spaces at different levels of coarsening, we used KaHyPar to generate 10000 random starting points, apply FM to each and stored these local optima. For each problem we then identified the (usually singleton) set of `quasi-global' optima. For each local optima, we measured its Hamming distance (and that of its inverse) to each of the global optima, and recorded the smallest distance (scaled [0,1]), together with the relative cut-size, i.e., divided by the landscape's estimated global minimum. This was done at $t=150$ and $t=15000$ for four hypergraphs from each of ISPD98, SPM, and SAT collections.

Landscapes were examined through a combination of visual analytics (scatter and kernel-density-estimate, KDE plots) and a model of the fitness-distance correlation (FDC). The FDC model is a linear regression of local optima $l$ in the form $cut(l_i) = m \times distance(l_i,g)$. The proportion of observed variation in relative cut-size that can be described by the model was recorded, i.e., the co-efficients of determination ($R^2$).
 
This analysis showed a significant similarity between problems, with the exception of {\it Stanford} where coarsening stops prematurely.  Fig.~\ref{fig:kde} shows KDE plots for the two thresholds overlaid with the FDC results for two typical hypergraphs. Note the $y$ scales were chosen to permit comparison between different thresholds and so significant numbers of local optima with high relative cut-sizes are not shown. This is why the linear regression lines lie above the main cloud of points visible at $t=150$. The results of this analysis, and the implications for search algorithm design are:

\begin{figure}[t]
    \includegraphics[width=0.5\columnwidth]{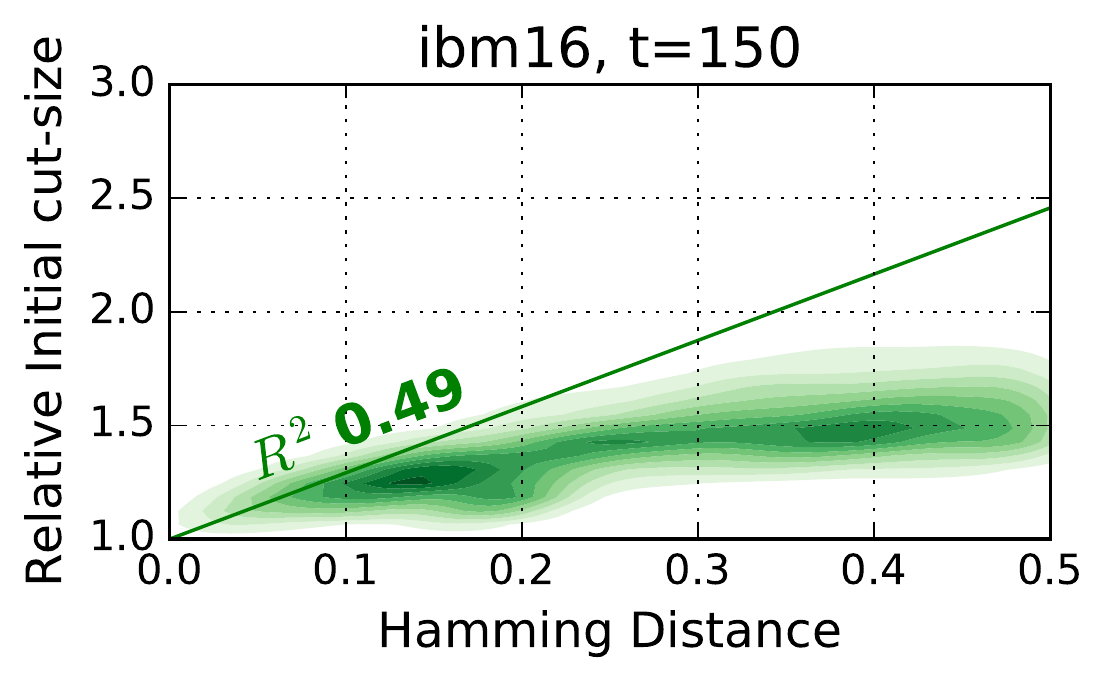}\includegraphics[width=0.5\columnwidth]{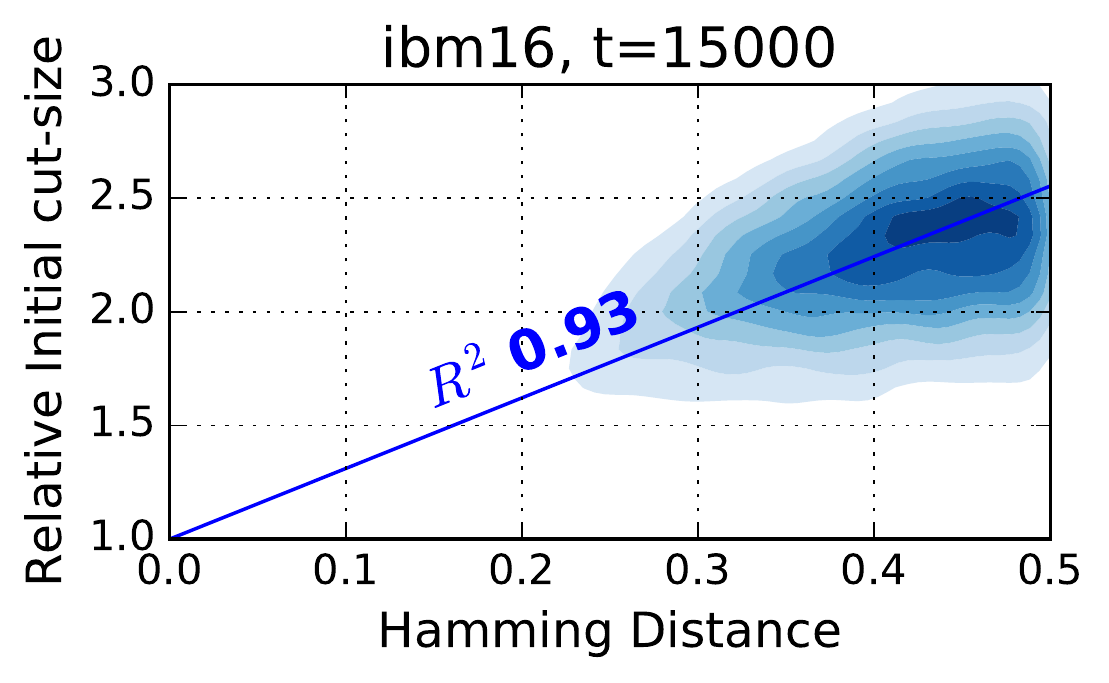}\\
    \includegraphics[width=0.5\columnwidth]{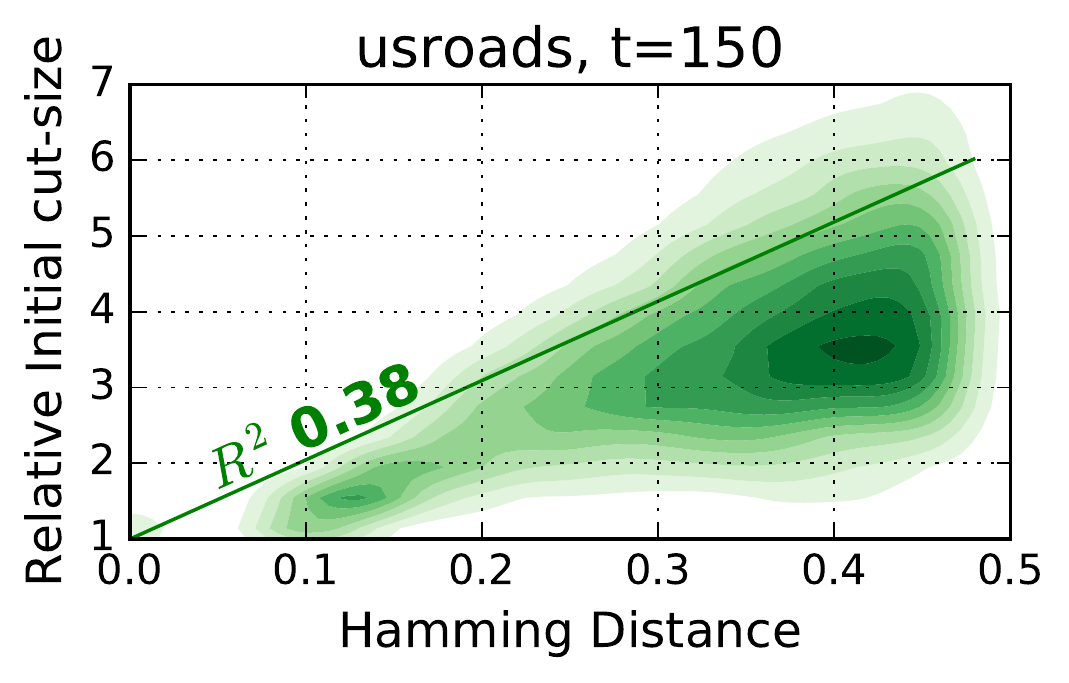}\includegraphics[width=0.5\columnwidth]{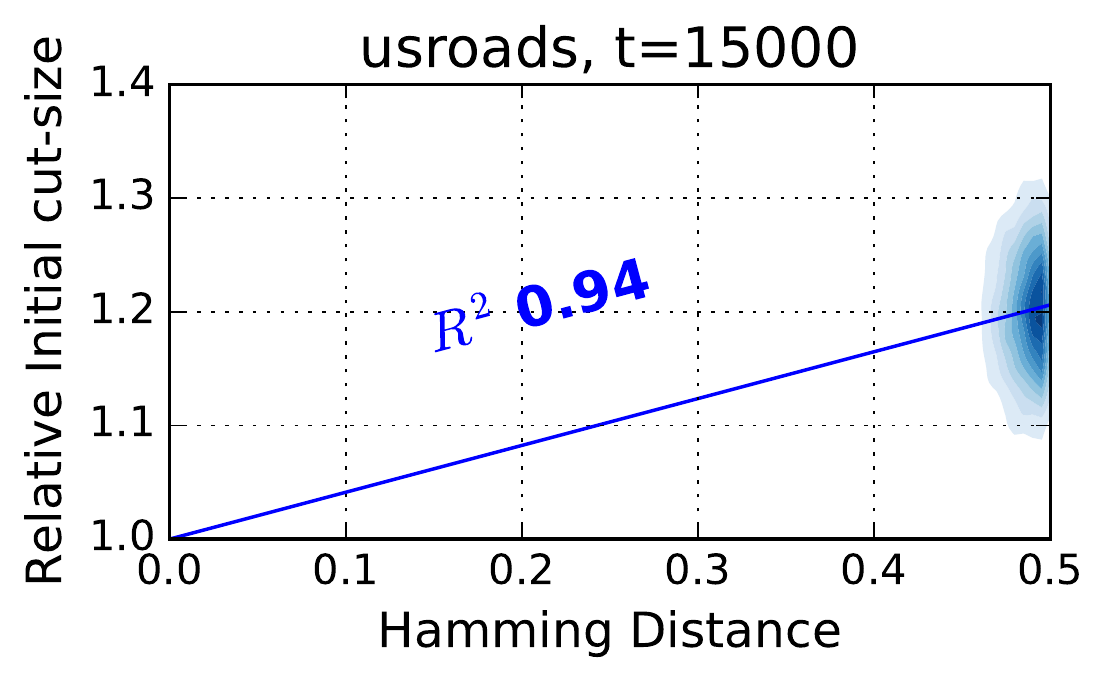}
    \caption{The relationship between local optima initial cut-size and Hamming distance at thresholds $t=150$ and $t=15000$. Each graph shows a kernel density plot of the results from 10000 randomly seeded FM local searches and FDC results. $y$-axes are scaled to facilitate comparison between thresholds and so do not show many poor optima for $t=150$.}
    \label{fig:kde}
\end{figure}

\begin{enumerate}
    \item On some problems the coarsening process was observed to stop prematurely, and at different values when repeated (e.g., between 34000 and 65000 hypernodes for \textit{Stanford}). This suggests that search algorithms should be designed to cope with large search spaces.
    \item The FM process greatly reduced cut-sizes and there was no correlation between the cut-sizes of solutions before and after improvement. This suggests a lack of global structure of the landscape as a whole, i.e., considering all points rather than just local optima. This indicates  algorithms should incorporate local search.
    \item All search landscapes contained large numbers of distinct local optima. Only a few tens of duplicates were found; more than one copy of the global optima was only found in 2 of the 24 runs, and never at $t=15000$. It was common to see cut-sizes an order of magnitude worse than the quasi-global optimum. This suggests that it is worth devoting computational effort to finding good starting points for the search process.
    \item On all landscapes there was a positive FDC, i.e., the global optimum was likely to be near other good local optimum. This mirrors previous findings on the related graph partitioning problem~\cite{Boese:1994, Merz:2000}. This suggests benefits for search algorithms that can exploit this information such as population-based search with some form of recombination.
    \item This effect was noticeably more present on the large landscapes ($t=15000$). This suggests that there may be a role for population-based search in partitioning at less coarse levels than is possible with single-member search algorithms such as BFS.
    \item There was almost always a `gap' between the best solution found and next best. The lack of duplicates makes it unlikely the global optima had large basins of attraction. Given the numbers of `good' local optima found just beyond this gap, this suggests a concentric structure. This may be because points ``in the gap'' are infeasible, or because the basins of attraction of the good-but-not-optimal local optima are large. Again this suggests a role for recombination, but as this has less effect as populations converge, it also suggests a changing role for mutation during search. Self-adaptation of mutation rates has often been shown successful in a wide range of domains~\cite{Meyer:2007} and simple approaches can be shown theoretically to be capable of overcoming both fitness and entropic barriers in combinatorial landscapes~\cite{Smith:2003}.
\end{enumerate}

\section{Sensitivity to EA design choices}
\label{sec:sensitivity}

\subsection{Population Seeding}

The landscape analysis suggests that for some hypergraphs there is good reason to devote significant effort to finding good starting points for search. To examine this hypothesis, and conversely, whether seeding is detrimental when those conditions do not apply, we exploit the portfolio of algorithms in the Pool as a  selection of heuristics for quickly finding approximate solutions. To examine the performance of the EA ($\mu=100, \lambda=1000$) with different amounts of initial seeding, experiments were run with the EA seeded with $\mu \times s$ Pool evaluations: for example, when $s=10$, the first 1000 evaluations are generated from the Pool before the EA begins.

\begin{figure*}[t]
    \centering
    \includegraphics[width=\columnwidth]{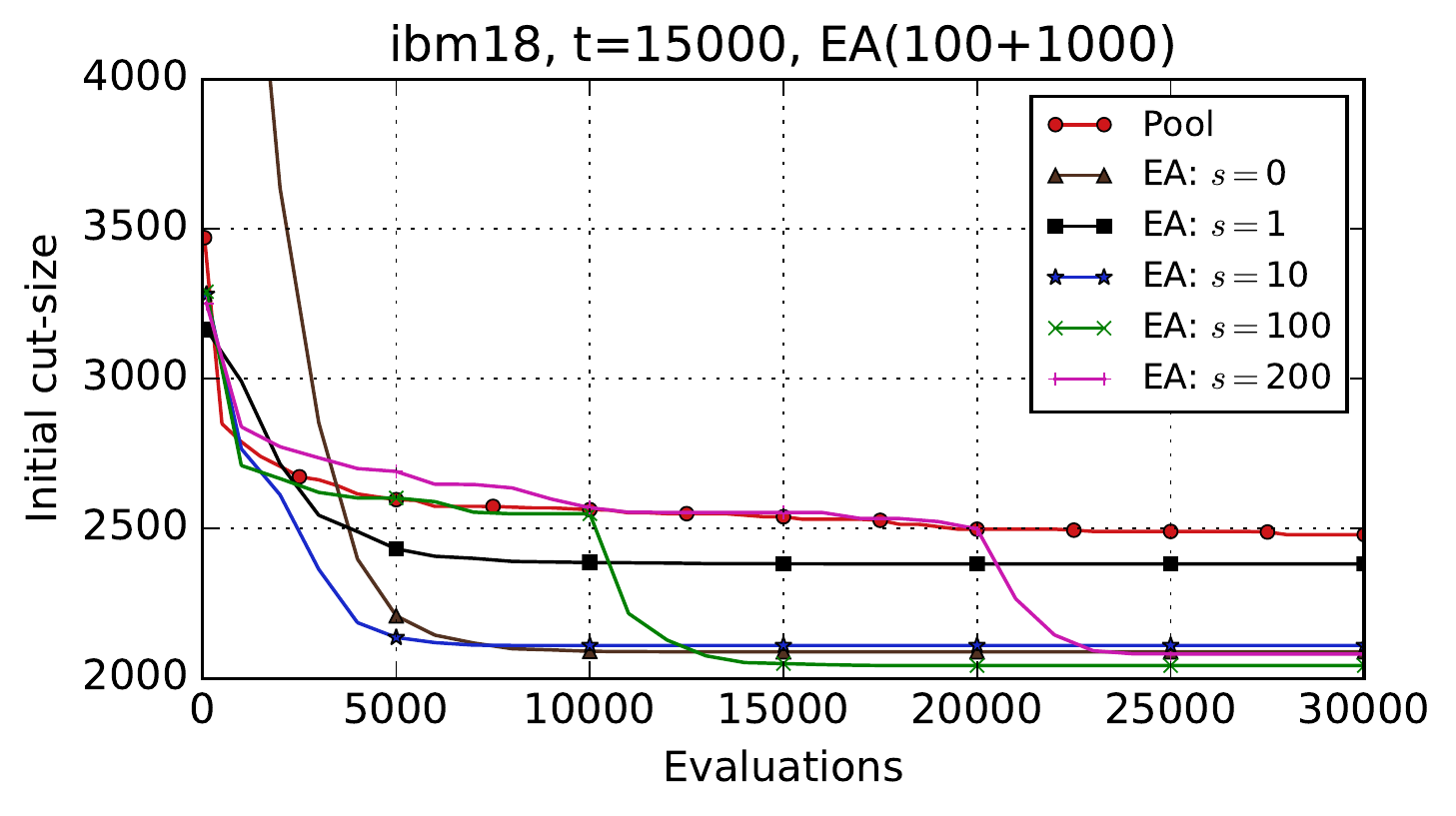}
    \includegraphics[width=\columnwidth]{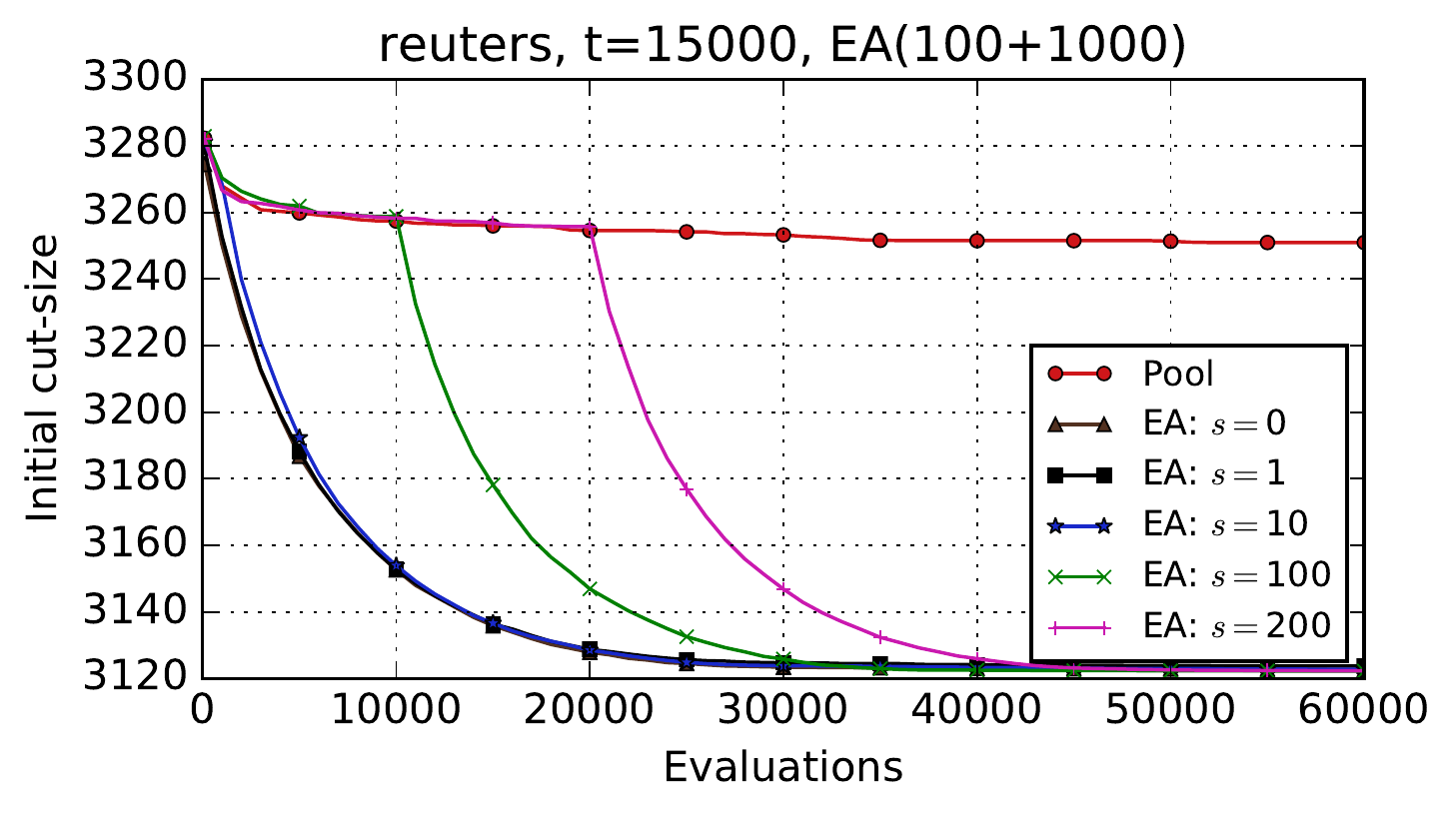}
    \includegraphics[width=\columnwidth]{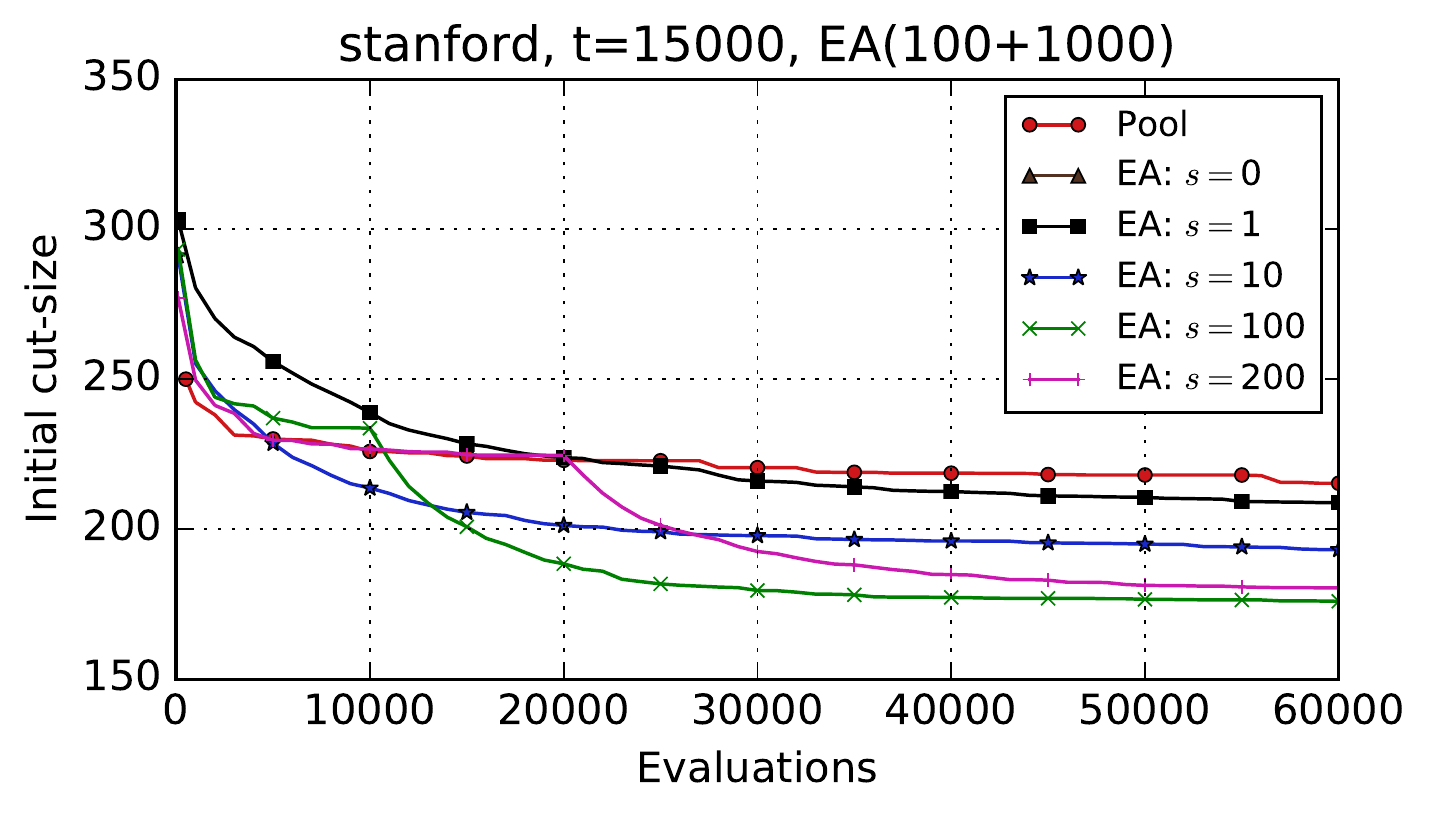}
    \includegraphics[width=\columnwidth]{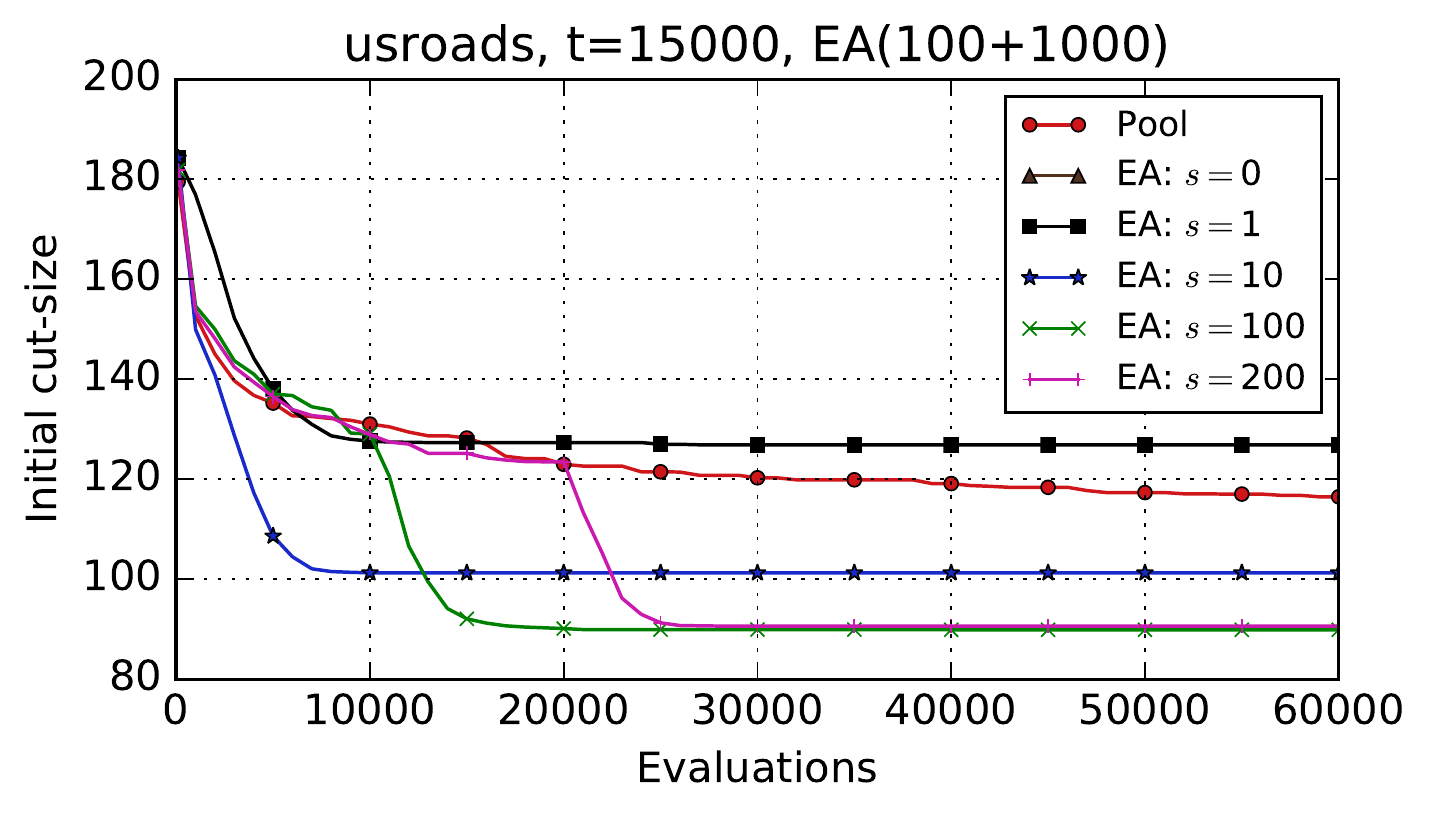}
	\caption{The affect of population seeding on the {\it ibm18, Reuters911, Stanford}, and {\it usroads} initial partitioning. Shown are the cut-sizes of the best solutions discovered by the Pool (circle), and the EA initially seeded with $\mu \times s$ number of Pool evaluations; $\mu=100$, $\lambda=1000$. On the {\it Stanford} and {\it usroads} hypergraphs the EA without seeding ($s=0$) is not observable since the cut-size values exceed the $y$-axis limit.}
    \label{fig:t15000-seeding}
\end{figure*}

In Fig.~\ref{fig:t15000-seeding} the cut-sizes of the best solutions discovered are shown for the \textit{ibm18}, \textit{Reuters911}, \textit{Stanford}, and \textit{usroads} hypergraphs at coarsening threshold $t=15000$. All results are averages of 20 runs. On both \textit{ibm18} and \textit{Reuters911}, the EA quickly identifies better solutions than the Pool algorithm regardless of the seeding strategy, showing that the evolutionary search is able to effectively follow a gradient in the fitness landscape. However, on \textit{Stanford} and \textit{usroads}, the EA without seeding ($s$ = 0) performs very poorly, being an order of magnitude worse than $s=100$ after 30000 evaluations. Given that so many local optima are present in such a fitness landscape, starting with fully random solutions ($s$ = 0) or only a few good solutions ($s$ = 1, $s$ = 10) can cause the EA to converge prematurely. Only by starting the EA at a suitable point in the landscape, here after 10000 Pool evaluations ($s$ = 100), is it able to consistently find very good solutions regardless of the effectiveness of coarsening. Further increasing the amount of seeding ($s$ = 200) did not result in additional improvements. In all following experiments therefore we use $s$ = 100, i.e., 10000 initial Pool evaluations.

The top-right KDE plot in Fig.~\ref{fig:kde} suggests a reason for these observations. The huge majority of local optima lie far from the global optimum and considering the high-density contours, there is little or no slope to guide the search towards the global optimum. Although there is a correlation between local optima cut-size and distance from the global optimum, this gradient only emerges when enough seeds have been considered to sample the lower-density contours of the KDE. 

\subsection{Population Size}
EA sensitivity to $\mu$ and $\lambda$ was explored by repeating the previous experiments across the spectrum of coarsening levels on the same 12 hypergraphs. A ratio of 1:10 was employed as this is a commonly used setting, especially with self-adaptive mutation~\cite{Serpell:2010}. The EA(10+100) was found to produce significantly worse final cut-sizes than EA(100+1000). However, EA(50+500) and EA(200+2000) were not significantly different than EA(100+1000). This shows that the EA is reasonably robust to these parameters and the use of 100+1000 is justified here for the use of fixed parameters. However, as shown in Table~\ref{table:tstar}, the optimum coarsening threshold $t^*$ differs for each hypergraph. Therefore, adaptive population sizing schemes would further optimise wall-clock partitioning time and have been shown to increase EA performance~\cite{Ali:2017}.

\subsection{Variation Operators}

Further experimentation on less coarsened hypergraphs ($t=15000$) confirmed results widely reported for graph partitioning~\cite{Kim:2011} that both the use of uniform crossover and parental alignment significantly improved performance. This finding remained consistent even with the use of self-adaptive mutation. For example, EA(100+1000) with $\mathcal{X}=80\%$ produced initial cut-sizes on average 30\% smaller than $\mathcal{X}=0\%$ on {\it ibm18} after 30000 evaluations, $p\le0.05$. 

Estimation of distribution algorithms (EDAs) have been used to generate many state-of-the-art results by replacing recombination and mutation with a process of building and then sampling probabilistic graphical models (PGMs) of the current populations. We adapted Pelikan's implementations of the Bayesian optimisation algorithm (BOA)~\cite{Pelikan:1999} to work within our seeding regime, and to explicitly exploit the representation's symmetry  during model building. With small $t$ no significant differences in performance were observed. However, the scalability of the model building process was an issue with large $t$. Runs on a MacBook Pro with a 2.8GHz 4-core Intel i7 processor with 16GB RAM were halted after 6 hours stuck in initial model building for both decision tree and graph-based variants of BOA, even after restricting the space of PGMs to bivariate models. Simplifying still further to a univariate model removed the ability to accurately capture interactions. Runs with $s$=100 initial seeding produced significantly larger mean initial cut-sizes after 30000 evaluations on the 4 hypergraphs in Fig.~\ref{fig:t15000-seeding}; 2422, 3154, 210, and 128 on {\it ibm18, Reuters911, Stanford} and {\it usroads}, respectively.

\subsection{Search at Different Coarsening Levels}

\begin{figure*}[!tbh]
    \centering
    \includegraphics[height=\graphheightscale\textheight, width=\columnwidth]{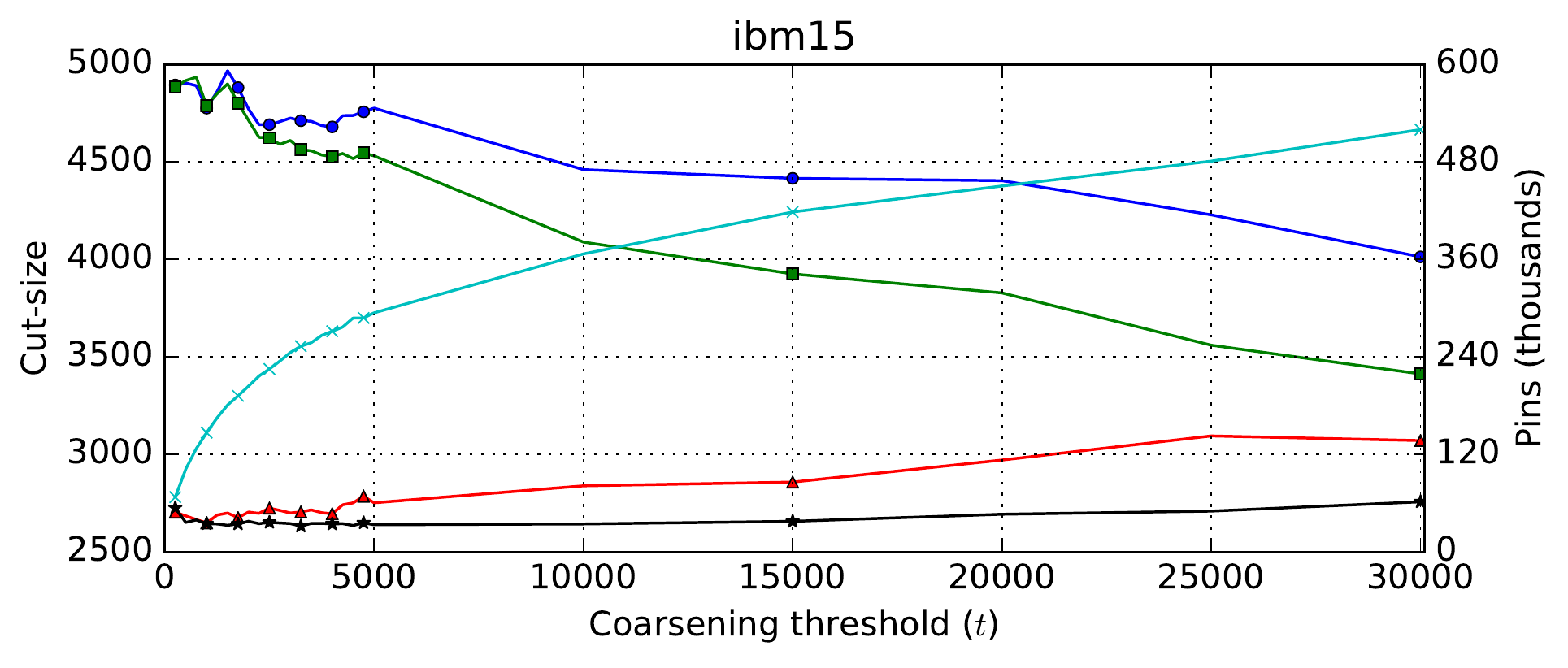}
    \includegraphics[height=\graphheightscale\textheight, width=\columnwidth]{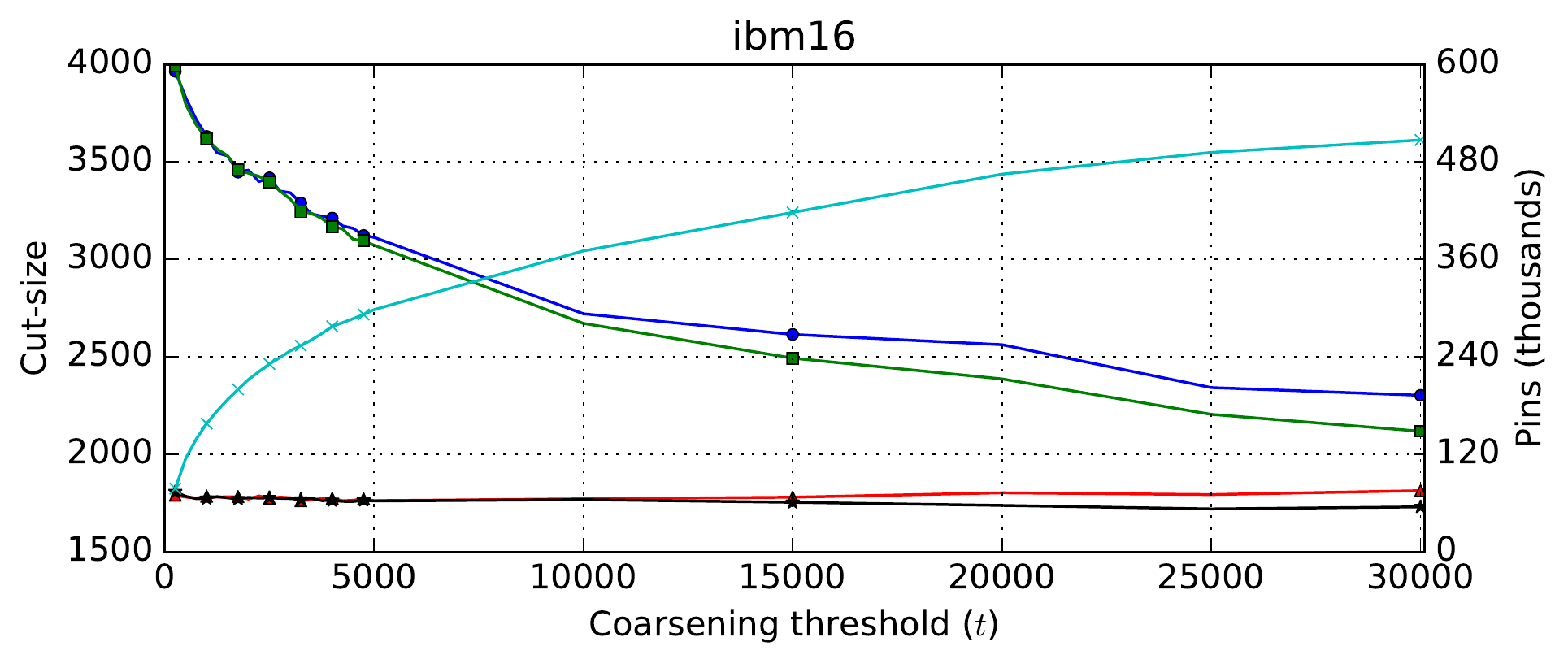}
    \includegraphics[height=\graphheightscale\textheight, width=\columnwidth]{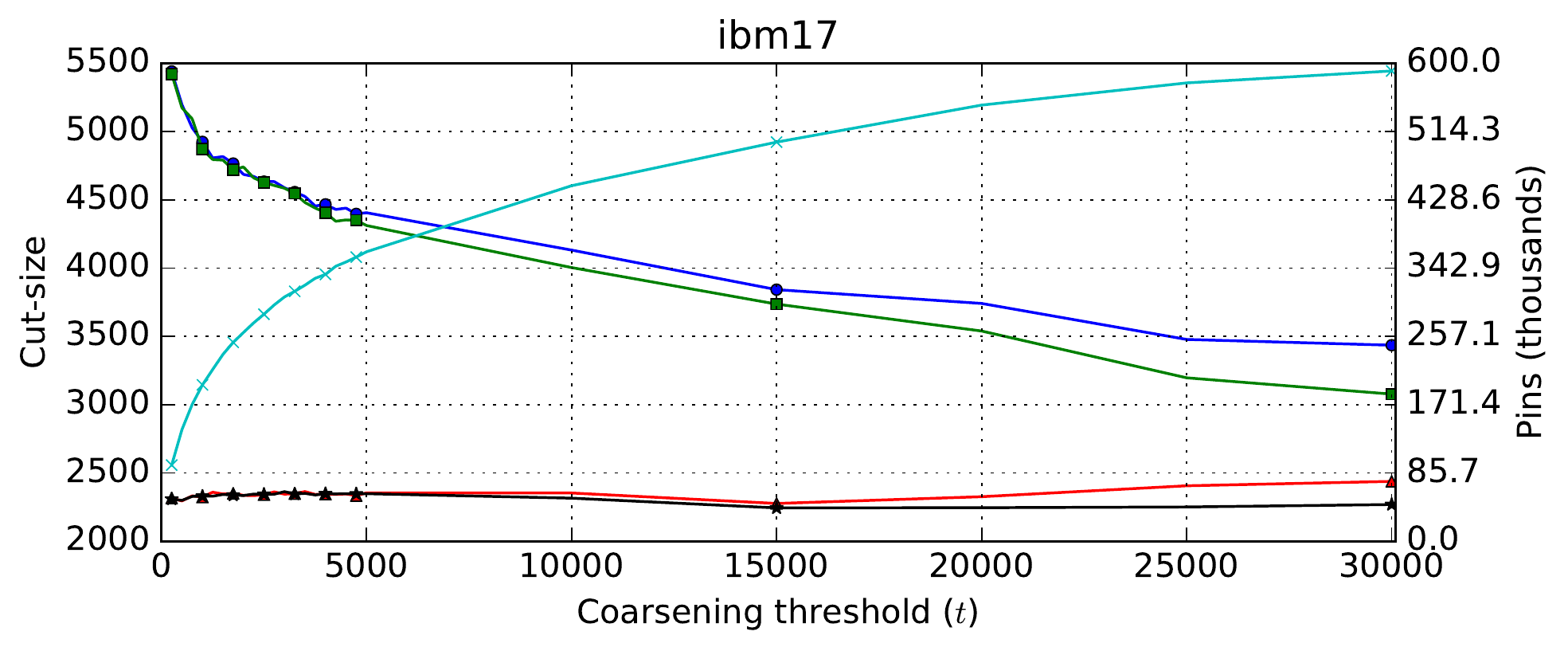}
    \includegraphics[height=\graphheightscale\textheight, width=\columnwidth]{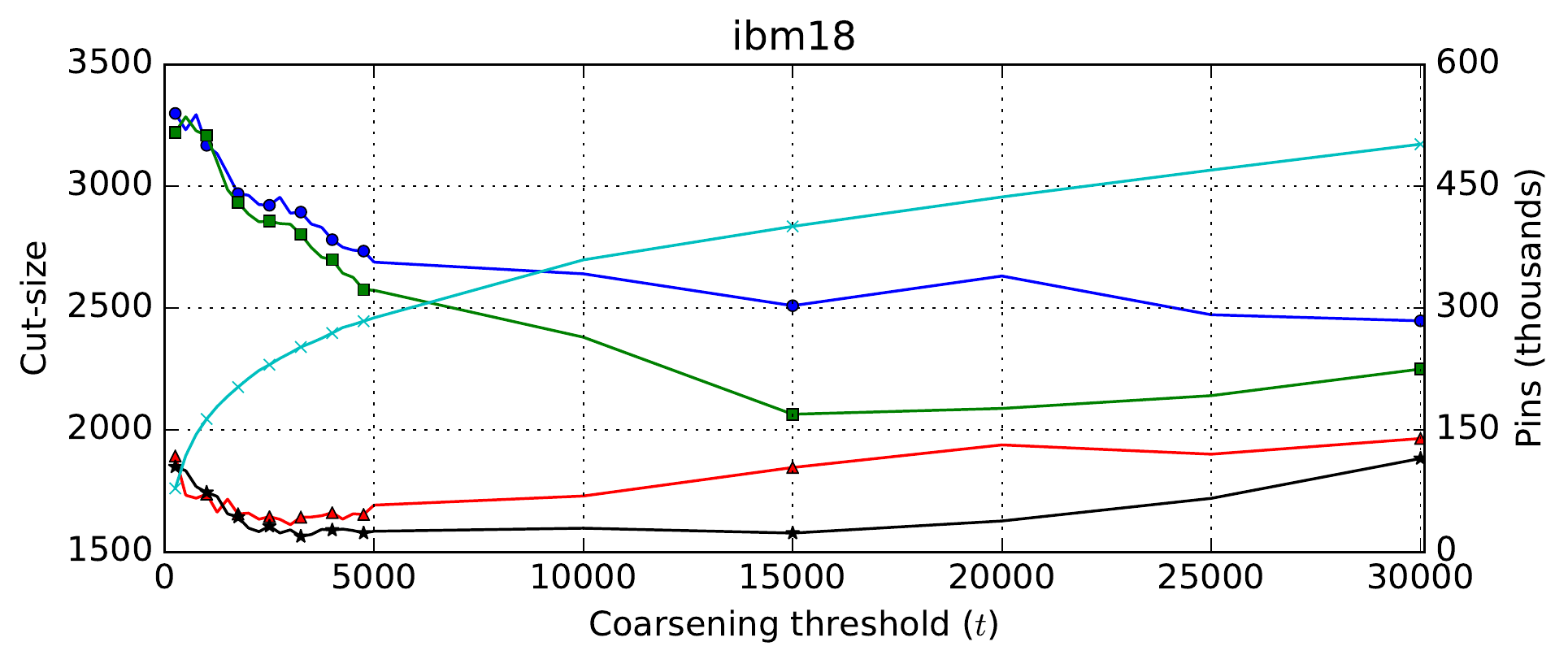}
    \includegraphics[height=\graphheightscale\textheight, width=\columnwidth]{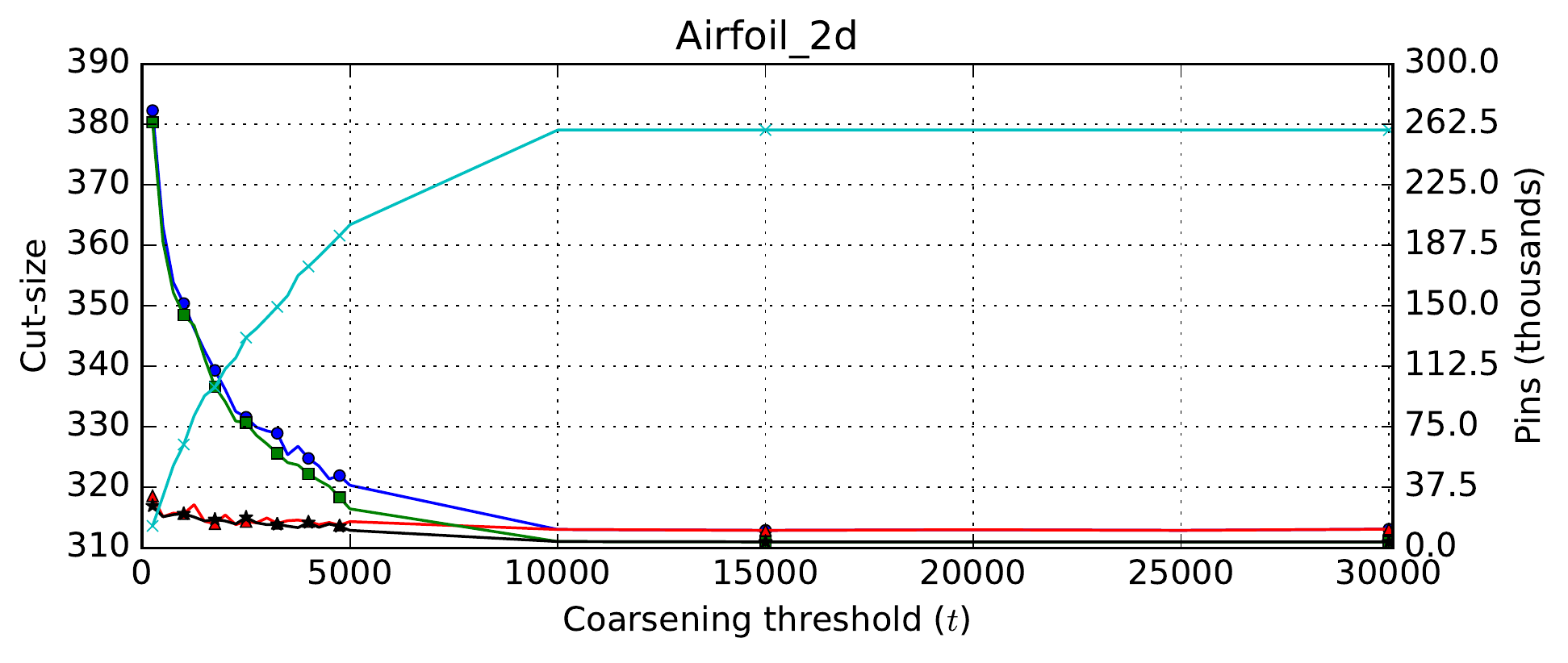}
    \includegraphics[height=\graphheightscale\textheight, width=\columnwidth]{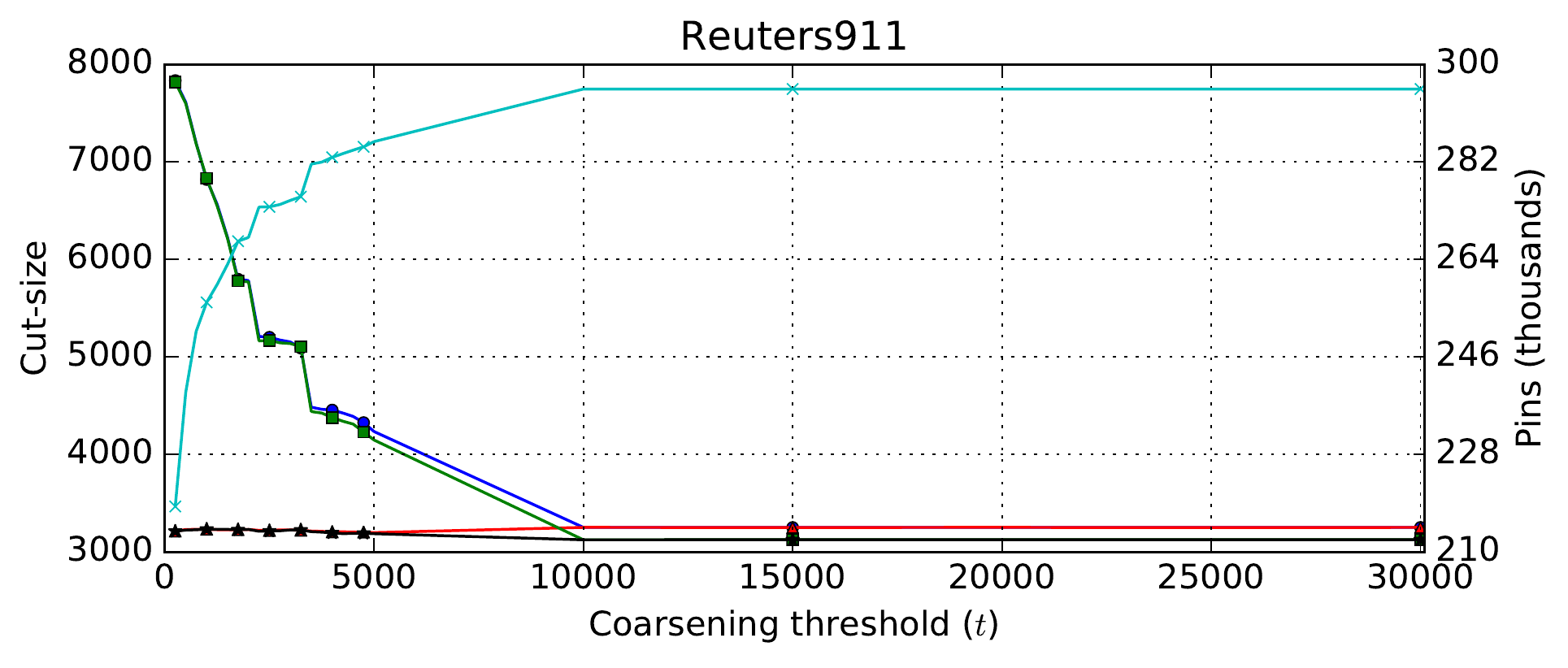}
    \includegraphics[height=\graphheightscale\textheight, width=\columnwidth]{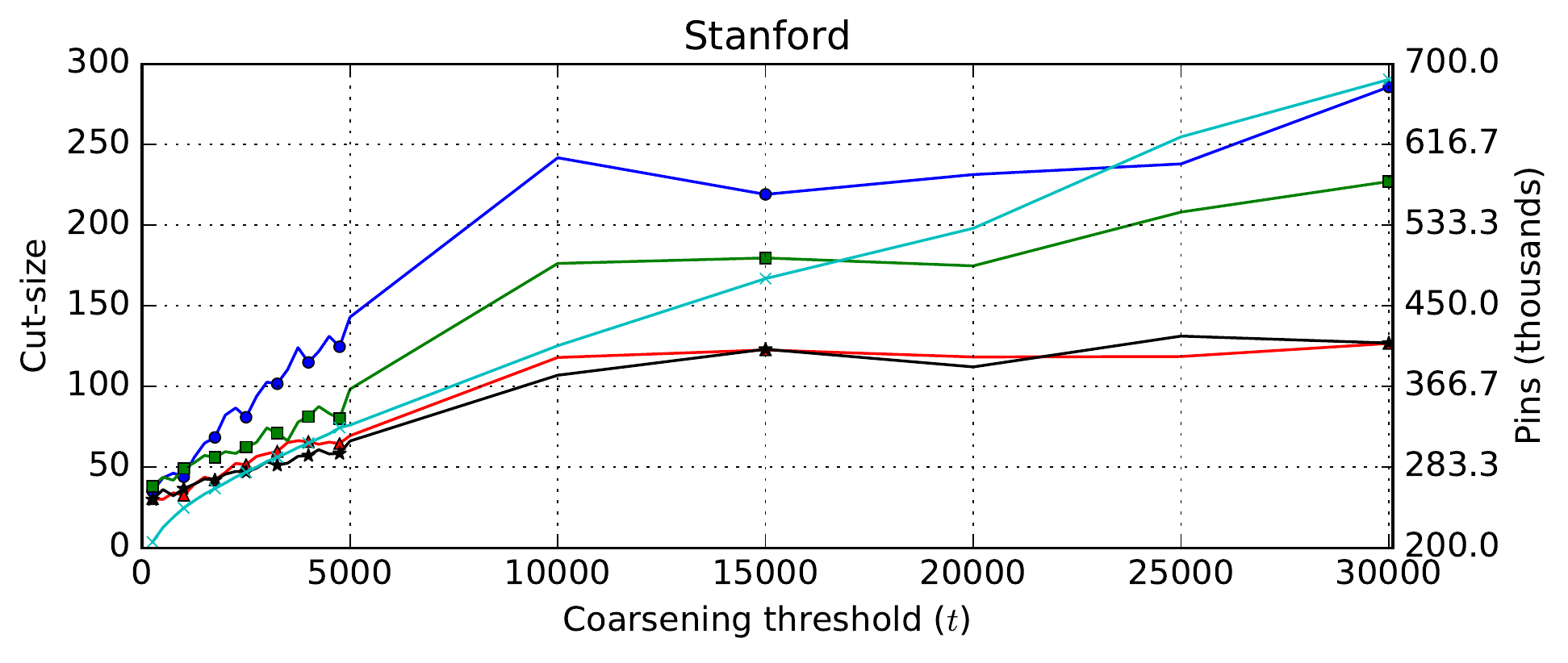}
    \includegraphics[height=\graphheightscale\textheight, width=\columnwidth]{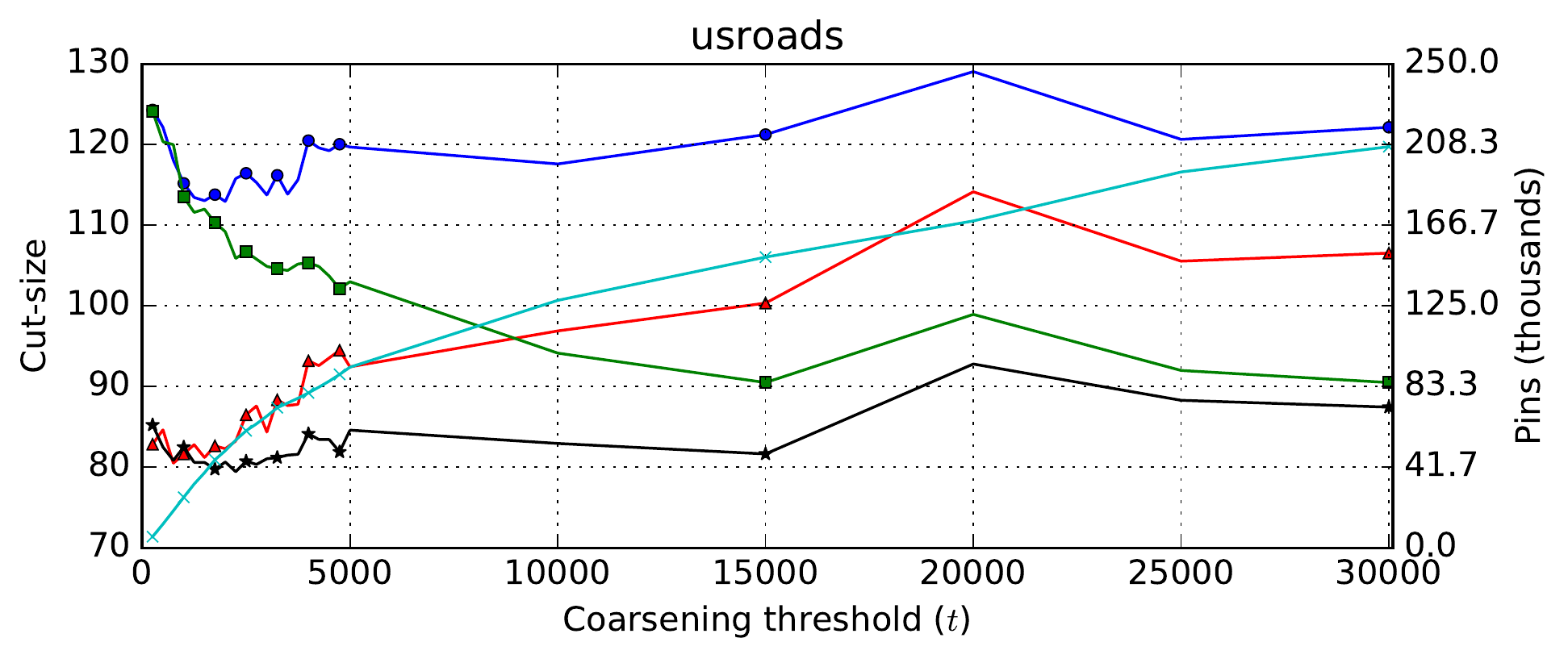}
    \includegraphics[height=\graphheightscale\textheight, width=\columnwidth]{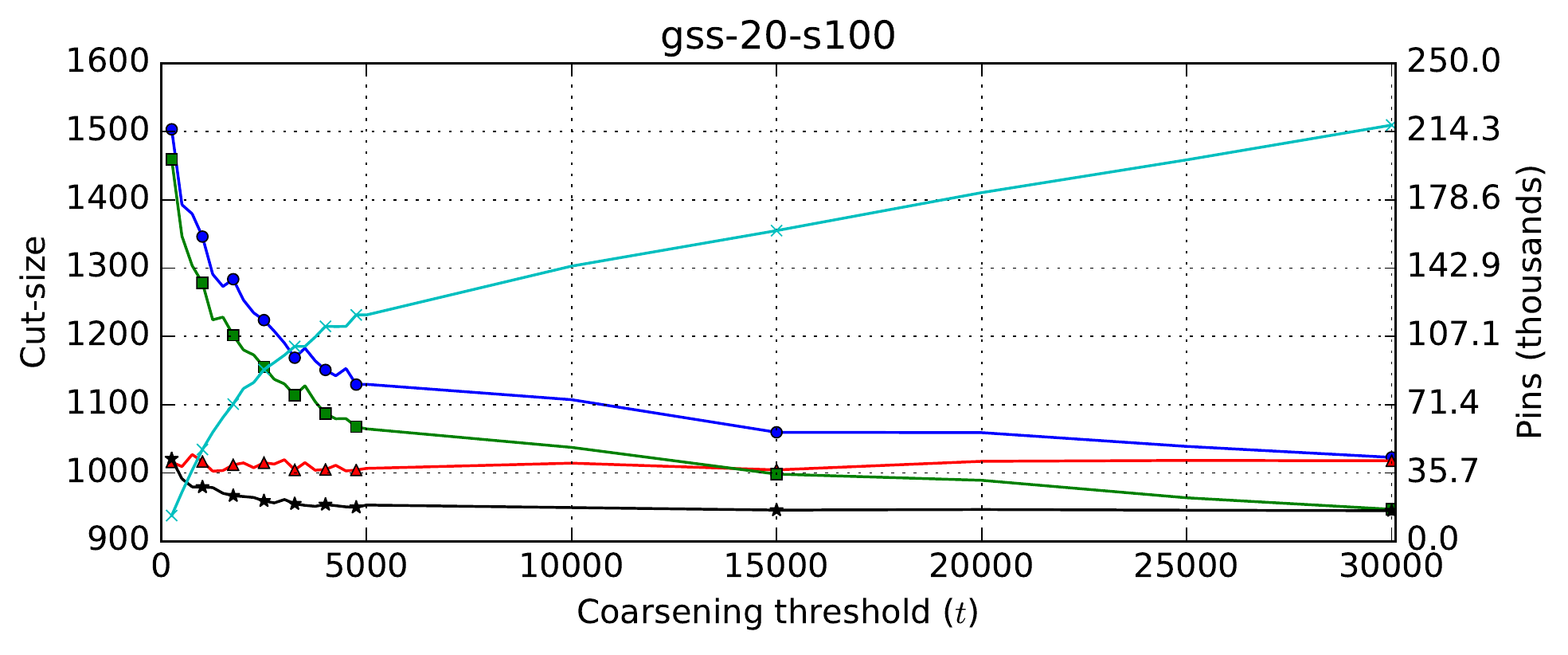}
    \includegraphics[height=\graphheightscale\textheight, width=\columnwidth]{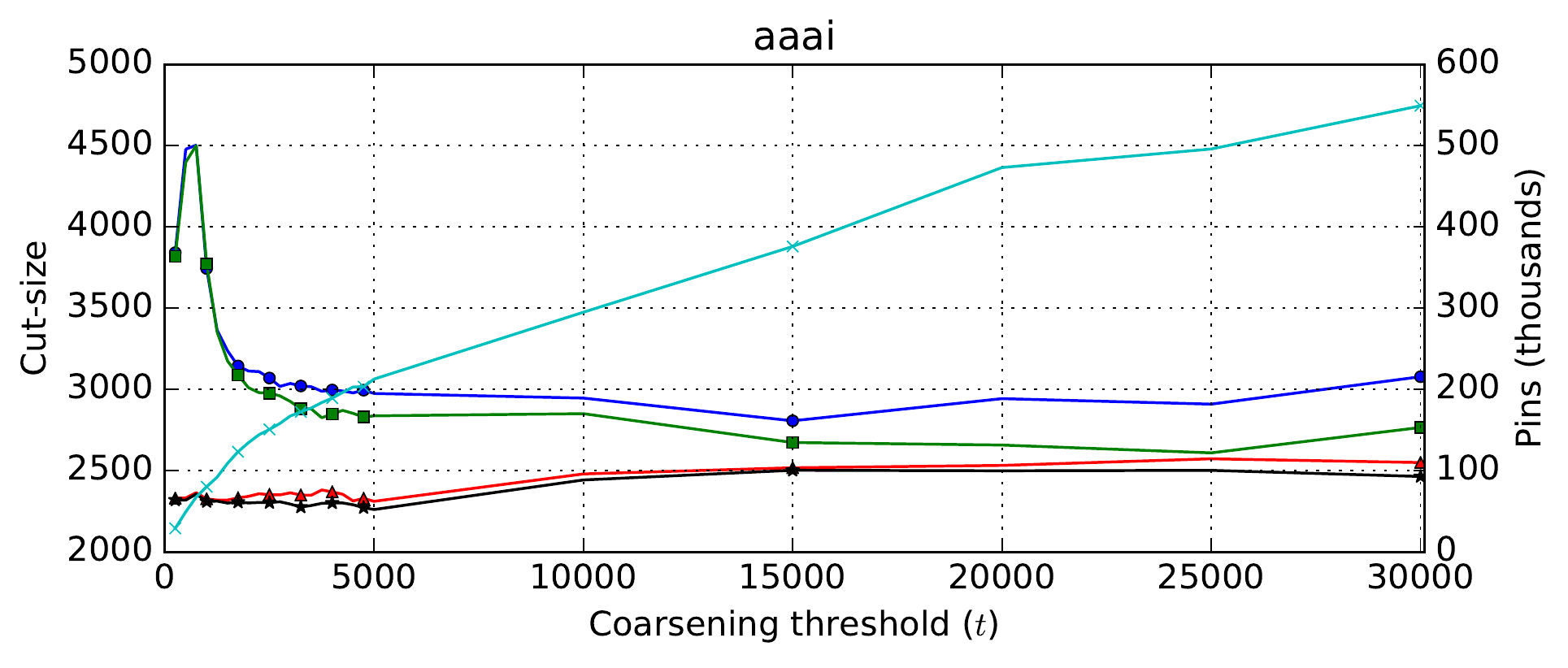}
    \includegraphics[height=\graphheightscale\textheight, width=\columnwidth]{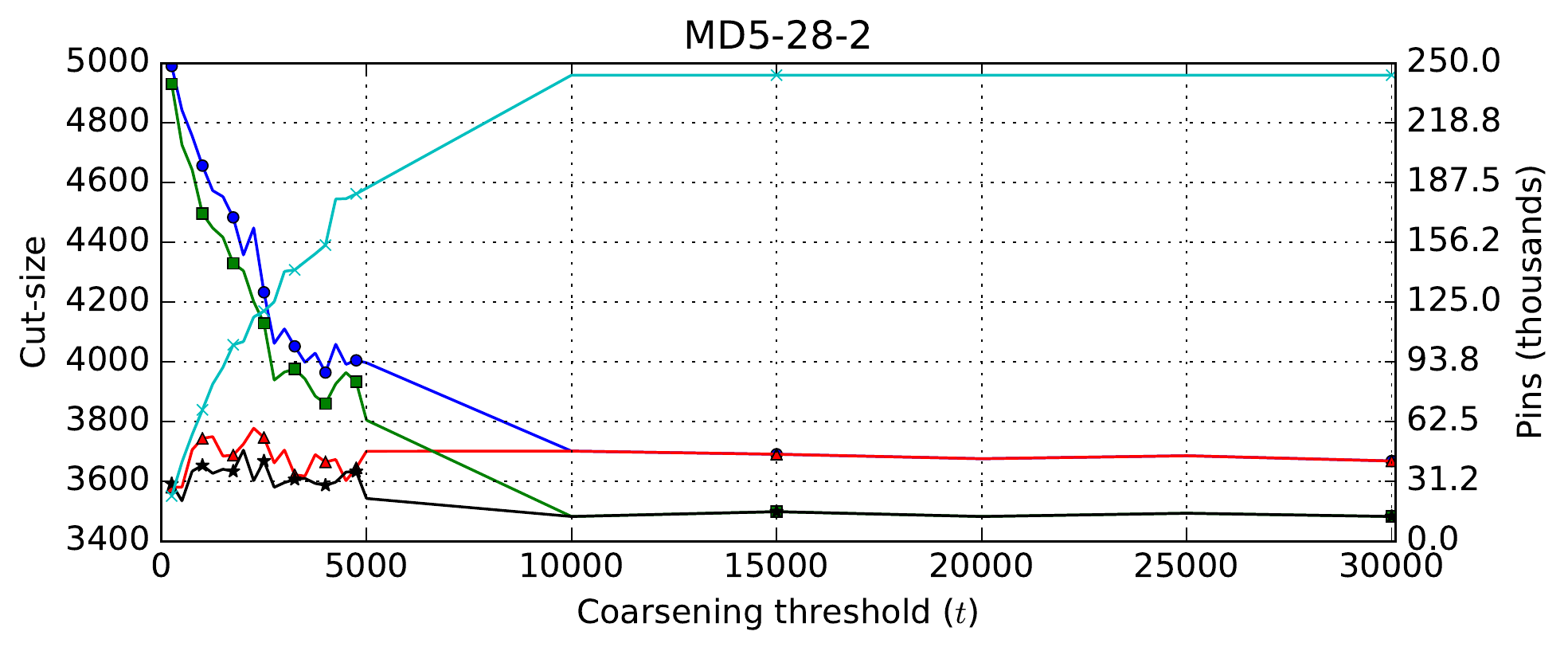}
    \includegraphics[height=\graphheightscale\textheight, width=\columnwidth]{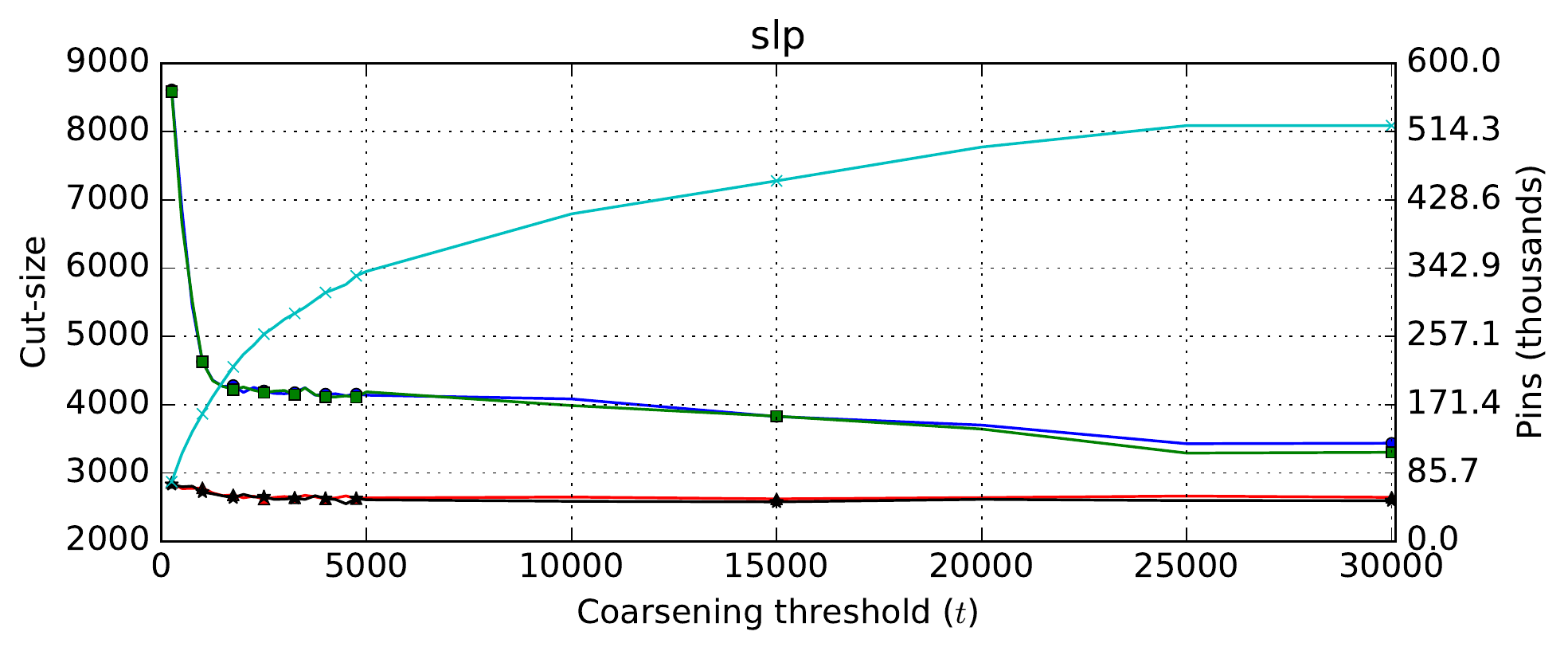}

	\caption{Cut-sizes for the initial and final partitioning of 12 hypergraphs from the ISPD98, Univeristy of Florida Sparse Matrix Collection, and 2014 SAT competition. Shown are the results of 20 runs of the Pool and EA(100+1000) run to 30000 evaluations at each coarsening threshold, sampled in intervals of 250 for $t\le5000$, and in intervals of 5000 for $t>5000$. Pool initial cut-size (circle); EA initial cut-size (square); Pool final cut-size (triangle); EA final cut-size (star); number of pins in the hypergraph (cross). For the {\it Airfoil\_2d, Reuters911}, and  {\it MD5-28-2} hypergraphs, $|V|\approx15000$, therefore the affects of coarsening can only be observed at $t<10000$.}
    \label{fig:spectrum}
\end{figure*}

The more coarsening performed on a hypergraph before partitioning, the more information is potentially hidden from the optimisation algorithm, i.e., it must move larger blocks. However, the less coarsening performed, the larger the search space and potentially the worse the optimisation algorithm will perform. To explore this relationship between algorithm and coarsening threshold, we examine the results of initial and final partitioning by the Pool and EA with $s$=100 seeding across a spectrum of coarsening levels. For each of the three classes of hypergraph, we perform experiments across the spectrum of coarsening thresholds on 4 of the 10 selected benchmark hypergraphs\footnote{ {\it ibm 15--18; gss-20-s100, aaai, MD5-28-2}, and {\it slp} from the SAT collection; and SPMs {\it Airfoil\_2d, Reuters911, Stanford}, and {\it usroads}.}. Additionally we ran tests at $t$ = 150 and $t$ = 15000 on all 30 hypergraphs. Results presented are an average of 20 runs of each algorithm run to 30000 initial partitioning evaluations at each coarsening threshold; each threshold is sampled in intervals of 250 for $t\le5000$, and of 5000 above that. The initial and final cut-sizes can be seen in Fig.~\ref{fig:spectrum}. 
    
\subsubsection{Overall Performance}

Using the AUC metric to compare performance across all coarsening thresholds, initial cut sizes found by the EA were smaller than those found by Pool on all 12 problems. The same is seen for final cut sizes with the exception of {\it Stanford}, where it should be noted that the coarsening algorithm produces hypergraphs with $|V|\ge30000$ (200000 pins) even at $t=150$.

\subsubsection{Highly Coarsened Hypergraphs}

The nature of the search landscapes for highly coarsened hypergraphs results in little difference between the algorithms. No statistically significant difference between algorithms was observed on any of the 30 benchmarks for either initial or final cut-sizes.

\subsubsection{Less Coarsened Hypergraphs}

The difference between algorithms becomes more significant the less coarsening is performed. For example, at $t=15000$ the EA mean best initial cut-sizes are significantly smaller than the Pool on all 10 of the ISPD98 hypergraphs (Wilcoxon rank-sum test, $p\le0.05$). Furthermore, these improvements in initial partitioning lead to smaller final cut-sizes. The mean and median are lower for the EA than the Pool algorithm on all 10 of the ISPD98 hypergraphs; but not significantly different at the 95\% confidence interval on \textit{ibm10} and \textit{ibm11}. On \textit{ibm18}, the EA mean inital and final cut-size were 20\% and 16\% smaller than the Pool.

Similar improvements to initial partitioning are found by the EA on the SPM hypergraphs. For example, with $t=15000$, the EA mean initial cut-sizes on 8 of the 10 SPM hypergraphs are significantly smaller than the Pool (Wilcoxon rank-sum test, $p\le0.05$); no significant difference was observed on the {\it nasarb} and {\it Andrews} hypergraphs. Interestingly, despite the improvement in initial partitioning, this only resulted in significant differences in final cut-sizes on the {\it Airfoil\_2d, Reuters911}, and {\it usroads} hypergraphs, where the EA resulted in improvements to mean final cut-size of 0.7\%, 4\%, and 15\% respectively. At this $t$ setting, no coarsening is performed on either the {\it Airfoil\_2d} or {\it Reuters911} hypergraphs and therefore the cut-sizes are entirely a result of the memetic EA.

For SAT hypergraphs at $t=15000$, both the mean EA initial and final cut-size is significantly smaller than the Pool on 6 of the hypergraphs ($p\le0.05$), with no significant difference on the other 4, again showing that the EA performs a more effective search on larger hypergraphs.

Performing Wilcoxon signed-ranks tests of the initial partitionings across all runs on the 10 ISPD98 hypergraphs confirms that the EA has a significantly lower cut-size than the Pool at $t=15000$ ($p\le0.05$). Moreover, this also translates to significant improvements in the final partitioning ($p\le0.05$). Similar results were found when repeating the class tests for the 10 SPM hypergraphs and the 10 SAT hypergraphs.

\subsubsection{Optimum Coarsened Hypergraphs}

Table~\ref{table:tstar} shows the smallest (average) final cut-sizes discovered by the Pool and EA across all coarsening thresholds on the 4 hypergraphs from each benchmark set. This shows that when the optimum coarsening threshold for each algorithm-problem combination is known, the smallest final cut-size discovered by the EA is less than the Pool algorithm on all 4 of the largest ISPD98 hypergraphs. On the SAT hypergraphs, the best EA final cut-sizes are on average smaller by 5.8\% on \textit{gss-20}, 2.2\% on \textit{aaai10}, 2.75\% on \textit{MD5-28-2}, and 2.6\% on \textit{slp-synthesis}. These improvements are statistically significant for all but \textit{ibm15} and \textit{Stanford}. The improvements were achieved by the EA carrying out a more effective search at the same or higher coarsening threshold than the Pool and therefore able to take advantage of any additional information in the larger initial hypergraph.

Also shown in Table~\ref{table:tstar} is the average total EA partitioning time, $time^*_{EA}$, relative to that taken by the Pool, $time^*_{Pool}$. As can be seen, the EA is faster on 7 of the 12 hypergraphs despite operating on a similar or larger initial hypergraph.

\begin{table}[t]
	\centering
	\caption{The smallest (average) EA and Pool final cut-sizes on four hypergraphs from each of the benchmark sets and the related coarsening thresholds. Cut-size highlighted in bold face where it is significantly different, $p\le0.05$.}
	\begin{tabular}{l l r r r r}
		\toprule
		Hypergraph & $t^*_{Pool}$ & $t^*_{EA}$ & $cut^*_{Pool}$ & $cut^*_{EA}$ &  $\frac{time^*_{EA}}{time^*_{Pool}}$ \\
		\midrule
		ibm15 & 1000 & 3250 & 2649 & 2632 & 2.69 \\
		ibm16 & 3250 & 25000 & 1762 & {\bf 1720} & 3.15  \\
		ibm17 & 15000 & 15000 & 2276 & {\bf 2244} & 0.74 \\
		ibm18 & 3000 & 3250 & 1612 & {\bf 1564} & 0.57 \\
		Airfoil\_2d & 15000 & 15000 & 312 & {\bf 311} & 0.66\\
		Reuters911 & 5000 & 10000 & 3199 & {\bf 3125} & 0.60 \\
		Stanford & 500 & 250 & 30 & 29 & 0.40 \\
		usroads  & 750 & 2250 & 80 & {\bf 79} & 1.87\\
		aaai10-planning  & 5000 & 5000 & 2312 & {\bf 2261} & 0.65\\
		gss-20-s100  & 1250 & 30000 & 1002 & {\bf 944} & 9.67 \\
		MD5-28-2 & 500 & 10000 & 3580 & {\bf 3483}  & 6.41\\
		slp-synthesis & 2500 & 4500 & 2618 & {\bf 2549} & 0.96\\
		\bottomrule
    \end{tabular}
    \label{table:tstar}
\end{table}

\subsubsection{Summary}

\begin{itemize}
    \item The results for all 30 hypergraphs at the coarsest level ($t$=150) show no significant difference between algorithms.
    \item However, with larger initial hypergraphs ($t$=15000), the EA significantly outperforms the Pool ($p\le0.05$).
    \item Furthermore, the wall-clock time of the Pool algorithm was significantly higher than the EA's ($p\le0.05$).
\end{itemize}

Moreover, results confirm our hypothesis that if initial partitioning is done on large hypergraphs, the picture changes dramatically. Taken as a whole, for the 12 instances where the spectrum of coarsening thresholds was explored:
\begin{itemize}
    \item The EA significantly outperforms the Pool algorithm over all coarsening thresholds (AUC metric).
    \item The final cut-sizes of the EA at $t^*$ are significantly smaller for all 12 hypergraphs than the Pool algorithm at the default $t$=150.
    \item Taking the optimum threshold for each algorithm-problem combination, and comparing the best-case cut-sizes across the 12 problems, the EA results are significantly better than the Pool algorithm ($p\le0.05$).
\end{itemize}

\section{Adaptive Coarsening to identify the EA niche}
\label{sec:adaptive}

The less coarsening is performed, the more information may be available to the initial partitioning algorithm to potentially achieve higher quality partitions. This is particularly evident in a number of the hypergraphs in Fig.~\ref{fig:spectrum} by observing the final cut-sizes where $t<5000$; see, for example, {\it ibm18}. However, for each algorithm there exists a point at which further increases in the size of the search space result in declining performance; for example, see the algorithm cut-sizes on the {\it ibm18} hypergraph where $t>15000$ in Fig.~\ref{fig:spectrum}. Simply selecting a fixed larger $t$ does not help since the `optimal' threshold is clearly hypergraph-dependent..

From Fig.~\ref{fig:spectrum} it can be seen that the sum of the number of vertices in each hyperedge, $|pins|$, initially declines relatively linearly with the number of hypernodes before reaching a point of exponential decay. This suggests that for each hypergraph there may exist a tipping point at the balance between maximal information content and maximal hypergraph compression, akin to `knee-points' in Pareto fronts. We therefore propose an adaptive coarsening scheme that halts hypernode contraction in response to  the changing characteristics of the hypergraph.

\subsection{Algorithm}

We perform a linear piecewise approximation of the curve based on a sliding window of observations, and seek to identify the knee-point at which the linear approximation is least representative of the curve. Coarsening occurs as normal until there are fewer than $t_{max} \times k$ hypernodes; here $t_{max}=15000$. Thereafter, a linear regression is performed on $|pins|$, sampled after every $t_s$ hypernodes have been contracted, and calculated on the most recent $t_n$ samples. Coarsening is terminated and initial hypergraph partitioning performed as usual when the correlation coefficient $R^2<t_r$ or the original $t=150$ threshold reached. See Algorithm~\ref{alg:ac}.

\begin{figure}[t]
    \removelatexerror
    \begin{algorithm}[H]
    	\SetNoFillComment
    	\small
    	\SetAlgoLined%
    	\DontPrintSemicolon
    	$R$: regression buffer of length $t_n$\;
    	\While{$|V| > t \times k$} {
    	    \For{each hypernode} {
    		    select contraction partner\;
    		    perform contraction\;
    		    \If{$|V| < t_{max} \times k$} {
    			    \If{$t_s$ hypernodes coarsened since last update} {
    				    update $R$ with $|pins|$\;
    				    $r \leftarrow$ coefficient of linear regression on $R$\;
    				    \If{$r^2 < t_r$} {
    					    stop coarsening\;
    				    }
    			    }
    		    }
    		    \If{$|V| < t \times k$} {
    			    stop coarsening\;
    		    }
    	    }
    	}
    	\caption{Adaptive coarsening stopping criteria}
    	\label{alg:ac}
    \end{algorithm}
\end{figure}
 
A grid search of these parameters was performed to minimise the final EA(100+1000) cut-sizes on the 12 hypergraphs for which partitioning was previously performed across the range of coarsening thresholds and the best performing parameters $t_s=50$, $t_n=100$ and $t_r=0.99$ were identified.

\subsection{Results}
Results show that over a wide range of different hypergraphs this simple adaptive threshold can identify better places to stop coarsening, although with some large variations:
\begin{itemize}
    \item Across all 30 hypergraphs there was an overall reduction in the mean final cut-size of 1.6\% ($p\le0.05$) compared with the results achieved at $t$=150; and a 1.25\% reduction ($p>0.05$) compared with results at $t$=15000. 
    \item The mean final cut-size is smaller on 22 of the 30 hypergraphs when using the adaptive threshold compared with the EA at $t$=150. This difference is statistically significant on 6 of the 10 ISPD98 hypergraphs, 2 of the 10 SPM hypergraphs (\textit{Reuters911} and \textit{usroads}) and 2 of the 10 SAT hypergraphs (\textit{gss-20-s100} and \textit{UCG-15-10p1}). Similar improvements are found when compared with the Pool at $t$=150.
    \item Excluding the 12 hypergraphs used for training the coarsening parameters, the EA achieves an overall reduction in the mean final cut-size of 1.8\% ($p\le0.05$) compared with the results achieved at $t$=150.
    \item Taken hypergraph-by-hypergraph, the mean final cut-size is smaller on 13 of the 18 hypergraphs. There is no significant difference compared with $t$=15000 and yet overall the average wall-clock time was $\approx 7.4\times$ faster.
    \item Total partitioning time with $t$=150 is of course much faster than the adaptively coarsened hypergraphs ($\approx10\times$), however with larger cut-sizes. Thus, showing the existence of the aforementioned knee-points. 
\end{itemize}

The use of a range of visual analytics tools failed to uncover any obvious relationships between the characteristics of the uncoarsened hypergraphs and the magnitude and direction of the performance difference arising from adaptive coarsening.

\section{Conclusions}
\label{sec:conclusions}
Our analysis of the state-of-the-art in hypergraph partitioning algorithms reveals that despite considerable sophistication, all algorithms use a somewhat arbitrary threshold for determining the size of the initial partitioning problem to be solved. This is perhaps driven by the poor scaleability of the search algorithms involved, such as BFS.

However, experimental analysis of the `searchability' of initial partition landscapes at different coarsening thresholds shows that larger landscapes may have properties that can be exploited by population-based search, and we derive some guidelines for algorithm design based on that analysis.

Experimental results confirm our hypothesis that there is valuable `niche' for EA-based search that leads to statistically significant reductions in final cut-size: up to 20\% compared to the default settings (Pool algorithm at $t$=150). Searching effectively in larger search spaces comes at a cost of approximately ten-fold in runtime, but this may well be warranted in many contexts such as `one-off' design, or where subsequent processing is needed within the partitions. 

Sensitivity analysis confirmed the guidelines derived from landscape analysis: recombination is useful, population size is not critical, and it is worth devoting a significant proportion of the computational budget to seeding the EA-base search. 

Examining the search performance of different algorithms at different coarsenening levels, we observe that there is a `sweet-spot' for EA-based search that is instance-dependent. We identify a novel, computationally cheap method for halting coarsening by monitoring the rate of change in information content as the hypergraph is contracted. This gives as good results as stopping at a predefined arbitrary larger threshold and with  runtimes reduced 7.5-fold.  

We do not claim to have developed the `best' EA to work in that niche. Rather, the aim of this paper was to establish the presence of a valuable role for EAs in hypergraph partitioning, working at a less coarsened level than currently used. In future work we will focus on (i) improved adaptive coarsening schemes, and (ii) tighter integration and re-use of information from the FM local search with the EA search processes and EDA model-building.

\section*{Acknowledgments}

The authors would like to thank the Karlsruhe Institute of Technology for KaHyPar and benchmark hypergraphs, and Martin Pelikan for his implementations of the BOA algorithm.



\end{document}